\documentclass[12pt]{article}  %

 \usepackage[letterpaper, left=1in, right=1in, top=1in, bottom=1in]{geometry}

\usepackage[colorlinks=true, linkcolor=blue!70!black, citecolor=blue!70!black]{hyperref}
\usepackage{microtype}

\usepackage{algorithm}

\usepackage[noend]{algpseudocode} 
 
\algblockdefx[class]{Class}{EndClass}[1]{\textbf{object} \textsc{#1}:}{\textbf{end object}}
\makeatletter  
\ifthenelse{\equal{\ALG@noend}{t}}%
  {\algtext*{EndClass}} 
  {}%
\makeatother

\usepackage{amsthm}
\usepackage{mathtools}
\usepackage{amsmath}
\usepackage{bbm}
\usepackage{amsfonts}
\usepackage{amssymb}

\usepackage{xpatch}
\usepackage{pifont}
\usepackage{array}
\usepackage{booktabs}
\usepackage{blindtext}
\usepackage{caption}

\usepackage{etoolbox}
\newtoggle{colm}
\togglefalse{colm}    
\newcommand{\arxiv}[1]{\iftoggle{colm}{}{#1}}
\newcommand{\colm}[1]{\iftoggle{colm}{#1}{}} 

\usepackage{microtype}
\usepackage{hyperref}
\usepackage{url}
\usepackage{amssymb}
\usepackage{booktabs}
\usepackage{physics}
\usepackage{xcolor} 
\usepackage{graphicx} 
\usepackage{comment} 
\usepackage{multirow}
\usepackage{enumitem}
\usepackage{subfigure}
\usepackage{longtable}

\newcommand{\refmodel}{\lmb} 

\arxiv{\date{}}

\arxiv{\usepackage{natbib}  
\usepackage{amsfonts,amssymb,mathrsfs}
\bibpunct{(}{)}{;}{a}{,}{,}
\usepackage{etoolbox} 
\apptocmd{\thebibliography}{\setlength{\itemsep}{0.5em}}{}{}  
}

\definecolor{darkblue}{rgb}{0, 0, 0.5}
\hypersetup{colorlinks=true, citecolor=darkblue, linkcolor=darkblue, urlcolor=darkblue}

\title{UCD: Unlearning in LLMs via Contrastive Decoding}

\colm{\author{Vinith Suriyakumar\\ 
MIT \\ 
\And
Ayush Sekhari$^*$ \\
MIT \\  
\And 
Ashia Wilson\thanks{~Equal Advisory Contribution.} \\  
MIT \\ 
}}

\arxiv{\author{Vinith Suriyakumar\\ 
{\normalsize \texttt{MIT}} \\ 
\and
Ayush Sekhari$^*$ \\
{\normalsize \texttt{Boston University}} \\  
\and 
Ashia Wilson\thanks{Equal Advisory Contribution.} \\ 
{\normalsize \texttt{MIT}} \\  
}}

\usepackage{amsmath}
\usepackage{amsthm}

\newtheorem{theorem}{Theorem}  
\newtheorem{prop}[theorem]{Proposition}  

\usepackage[suppress]{color-edits}    
\addauthor{as}{purple} 
\addauthor{vms}{blue} 

\newcommand{\cY}{\mathcal Y} 

\usepackage{bm}

\usepackage{multirow,nicefrac}

\usepackage[most]{tcolorbox} 
\PassOptionsToPackage{table}{xcolor}
\usepackage{xcolor}
\usepackage{colortbl}

\definecolor{lightgray}{gray}{0.9}  %
\definecolor{llamacol}{rgb}{0.19, 0.55, 0.91}
\definecolor{mistralcol}{rgb}{0.93, 0.53, 0.18}
\definecolor{phicol}{rgb}{0.31, 0.78, 0.47}

\usepackage{longtable}

\tikzset{
  dotted_line/.style={dash pattern=on 1pt off 4pt, line width=0.5pt}
}

\newcommand{\smg}{\mathrm{A}_{\mathrm{clean}}}
\newcommand{\smb}{\mathrm{A}_{\mathrm{corr}}} 
\newcommand{\lmg}{\mathrm{P}_{\mathrm{clean}}}

\newcommand{\lma}{\mathrm{P}_{\mathrm{aligned}}}

\newcommand{\lmb}{\mathrm{P}_{\mathrm{corr}}}

\begin{document}

\colm{
\ifcolmsubmission 
\linenumbers
\fi
}

\maketitle

\begin{abstract}
Machine unlearning aims to remove specific information, e.g. sensitive or undesirable content, from large language models (LLMs) while preserving overall performance. We propose an inference-time unlearning algorithm that uses contrastive decoding, leveraging two auxiliary smaller models, one trained without the forget set and one trained with it, to guide the outputs of the original model using their difference during inference. Our strategy substantially improves the tradeoff between unlearning effectiveness and model utility. We evaluate our approach on two unlearning benchmarks, TOFU and MUSE. Results show notable gains in both forget quality and retained performance in comparison to prior approaches, suggesting that incorporating contrastive decoding can offer an efficient, practical avenue for unlearning concepts in large-scale models. 

\end{abstract}

\ascomment{Things to do for Arxiv:
~ \\ Layer 1: 
\begin{enumerate}
\item Understand whether our method can be used with compositions.
    \item Verify the approximation assumptions to justify the method.
\end{enumerate}
~\\ Layer 2: 
\begin{enumerate}
    \item Experiments on \(\phi\)-models.  
    \item Any other ablations to add? 
    \item Can we also experiment with "in-context unlearning" as a benchmark? 
\end{enumerate}
~\\ Layer 3 (Future work / Another paper):  
\begin{enumerate}
    \item Anything with concepts? At the very least, provide a discussion of "concept unlearning" in the discussion. ~\\
    "What is a concept?" 
    \end{enumerate}
}
\ascomment{~\\ Next steps: Inference time unlearning:
\begin{enumerate}
    \item Not yet applied to unlearning. 
\end{enumerate}}

\section{Introduction}

Large Language Models (LLMs) achieve impressive general capabilities thanks to massive training datasets and compute.  However, these capabilities raise significant safety and security concerns, including copyright violations~\citep{karamolegkou2023copyright}, harmful content generation, and retention of dangerous knowledge (e.g., bioweapon instructions)~\citep{shevlane2023model}.  Retraining models to address these issues by excluding problematic data is impractical at scale. This has led to growing interest in efficient methods for {\em machine unlearning}, which aim to remove specific information from trained models without   retraining.

The field of machine unlearning began with a focus on removing the influence of specific training data points from trained machine learning models~\citep{cao2015towards, bourtoule2021machine, neel2021descent,  sekhari2021remember, ghazi2023ticketed, suriyakumar2022algorithms}.  This initial motivation arose primarily from compliance with emerging privacy regulations, such as the EU's General Data Protection Regulation (GDPR)~\citep{voigt2017eu} and the California Consumer Privacy Act (CCPA)~\citep{CCPA2018}, both of which enforce the \textit{Right to be Forgotten}. More recently, researchers concerned with AI safety have broadened the scope of machine unlearning to also include removing unwanted or harmful knowledge from large language models~\citep{li2024wmdp, barez2025open, zhang2023forget}.

So far, two broad classes of unlearning algorithms have been proposed for LLM unlearning: {\em finetuning-based} approaches and {\em representation-engineering} approaches. Finetuning algorithms define an objective to represent ``unlearning" and optimize it using samples of data to be forgotten (i.e., the {\em forget set}). The canonical example is gradient ascent, which maximizes empirical loss on the forget set. Extensions of this approach incorporate additional loss terms to maintain model utility~\citep{jang2022knowledge, yao2023large, chen2023unlearn, schwarzschild2024rethinking} or modify alignment procedures, such as direct preference optimization~\citep{rafailov2024direct}. Representation-engineering methods propose objectives to modify internal representations of the model with respect to the forget set, typically by projecting them onto random or orthogonal subspaces~\citep{li2024wmdp}. A shortcoming of both of these classes of methods is that they are expensive to run and suffer from poor forget-utility tradeoffs~\citep{shi2024muse}.

\begin{figure}[t]
\centering
\includegraphics[width=0.9\textwidth]{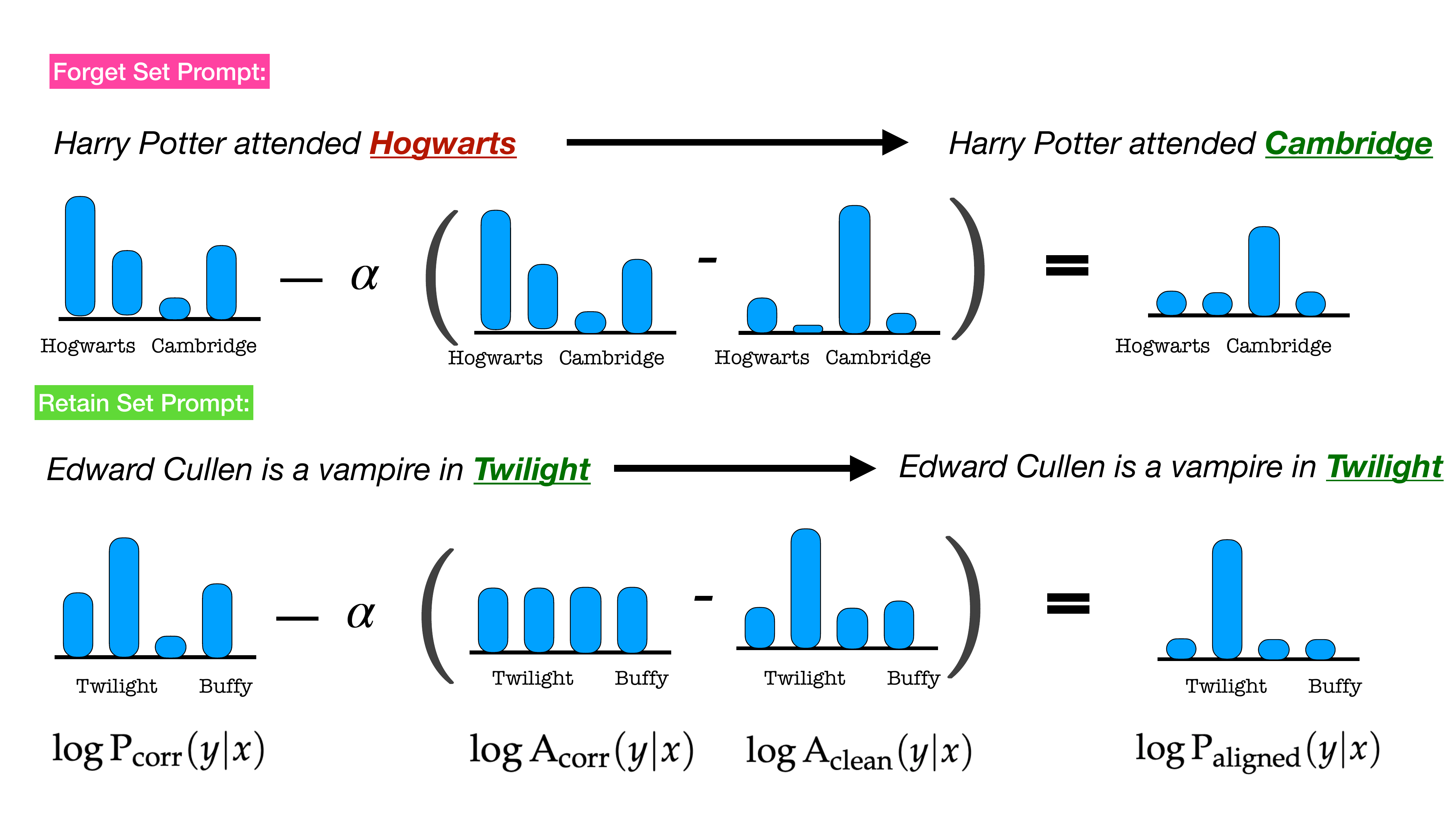}
\caption{Illustration of contrastive decoding at inference time in UCD. In the top row, we prompt the model with a sentence from our forget set corpus. The first distribution represents the original model we would like to unlearn from. The difference between our two auxiliary models guides the distribution to suppress the information related to Harry Potter. Meanwhile, in the second row, on a prompt we would like to retain the difference remains small leaving the output unaffected.}  
\label{fig:placeholder}
\end{figure}

Motivated by recent advances in inference-time methods that improve reasoning and alignment without extensive retraining, we propose \textbf{Unlearning via Contrastive Decoding (UCD)}, a novel inference-time unlearning algorithm inspired by contrastive decoding principles~\citep{li2023contrastive}. UCD leverages two small auxiliary models, one trained exclusively on the forget set and another trained on the retain set, allowing it to effectively remove undesirable knowledge at inference (Figure~\ref{fig:placeholder}). This approach significantly improves the forget-utility tradeoff and sets new state-of-the-art benchmarks on established unlearning datasets (TOFU and MUSE News). Additionally, due to its computational efficiency, UCD enables practical unlearning even on extremely large models such as Llama2-70B, a task previously infeasible with existing approaches. \textbf{Our main contributions are:}

\begin{itemize}[leftmargin=*]
\item \textbf{Contrastive Decoding-Based Unlearning:} We introduce UCD, an efficient inference-time unlearning algorithm utilizing two auxiliary models trained separately on forget and retain data. Whenever it is possible to obtain a clean model trained solely on the forget set, or when the data is sufficiently separable to allow targeted fine-tuning, UCD can be easily applied. 

\item \textbf{Superior Forget-Utility Tradeoff:} UCD significantly outperforms existing methods on standard machine unlearning benchmarks (TOFU, MUSE), achieving forget performance indistinguishable from retraining and improved utility due to contrastive decoding's enhanced text quality. 

\item \textbf{Scalability to Significantly Larger Models:} Unlike existing weight-modifying unlearning methods constrained by computational costs, UCD demonstrates practical inference-time unlearning on significantly larger models, including Llama2-13B and Llama2-70B, only requiring 2 L40s for unlearning on Llama2-13B and  4 NVIDIA H200s for unlearning on Llama2-70B. Whereas all pre-existing baselines required at least 2 A100s for unlearning on Llama2-13B and are infeasible on Llama2-70B on 8 H200s.

\end{itemize}

\section{Background and Related Work}

\paragraph{Machine unlearning.} Our work builds on a growing body of research on machine unlearning~\citep{bourtoule2021machine, nguyen2022survey, cao2015towards, gupta2021adaptive, suriyakumar2022algorithms, sekhari2021remember, ghazi2023ticketed, kurmanji2023towards, lev2024faster, lucki2024adversarial}, which aims to develop methods that efficiently modify trained machine learning models to forget specific portions of their training data. In the case of classical discriminative models, the motivation often stems from privacy concerns, particularly the need to protect individuals whose data may have been used during training. A major driver behind this line of research was the introduction of Article 17 of the European Union’s General Data Protection Regulation (GDPR), which codifies an individual’s ``right to be forgotten''\citep{gdpr2016}. Various other legislations have followed including California
Consumer Privacy Act (CCPA),  Canada’s proposed Consumer Privacy Protection Act (CPPA)), and more recently in Australia \citep{karp2023australia}.  More recently, the scope of machine unlearning has expanded to include modern generative AI models, which pose additional challenges such as the potential reproduction of copyrighted material, generation of harmful or explicit content, and leakage of sensitive training data~\citep{zhang2023text, carlini2021extracting}. 

\paragraph{Unlearning and alignment in LLMs.} Machine unlearning for Large Language Models (LLMs) has emerged as a rapidly growing area of research~\citep{liu2024large,jang2022knowledge,kumar2022privacy,zhang2023forget,pawelczyk2023context,yao2023large,zhang2024negative,wang2024large,jia2024soul,lu2022quark,liu2024rethinking,ishibashi2023knowledge,thaker2024guardrail,kadhe2024split,fan2024simplicity}. Given the inherent difficulty of exact unlearning, most existing approaches rely on approximate methods such as fine-tuning and representation engineering~\citep{yao2023large,eldan2023whos,jia2024soul,zhang2024negative,li2024wmdp, ilharco2022editing,liu2022continual} or prompt-based and in-context learning techniques~\citep{thaker2024guardrail,pawelczyk2023context,liu2024large}. Numerous benchmarks and evaluations have been developed to measure the effectiveness of these heuristical unlearning algorithms~\citep{maini2024tofu, shi2024muse, li2024wmdp}. We also highlight that test-time methods have started to gain prominence in LLM alignment, specifically using token-level rewards to guide generations~\citep{xu2024genarm}. We view our work as a similar family of methods where UCD provides a new reward designed for unlearning and representing the next token distribution if the model was trained without the forget set.

We defer a detailed description of the baselines that we compare to in our experiments, as well as additional related work on model-editing for unlearning, to Appendix \ref{app:background}.


\section{UCD: Unlearning via Contrastive Decoding}
This work focuses on the problem of \textit{machine unlearning} for large language models (LLMs). Given an initial model $\refmodel(y|x)$, referred to as the \textit{corrupted} or \textit{reference model}, 
that has been trained on a dataset $\mathcal{D} = {(x_i, y_i)}_{i=1}^{n}$, the central objective of machine unlearning is to effectively erase all information related to a designated subset of the dataset $\bm{\mathcal{D}_{\texttt{\bf forget}}} \subseteq \mathcal{D} $, termed the \textit{forget set}, while preserving the model's performance on the remaining subset $\bm{\mathcal{D}_{\texttt{\bf retain}}}$, known as the \textit{retain set}.

We approach the unlearning problem by leveraging auxiliary models to adjust the sampling distribution of the reference model. In particular, suppose there exists some public dataset $\mathcal{D}_{\texttt{pretrain}}$ that does not contain $\bm{\mathcal{D}_{\texttt{\bf forget}}}$, and a clean base model \(A\)  trained on $\mathcal{D}_{\texttt{pretrain}}$. Using this base model, we will first train two auxiliary models  $\smb$ and $\smg$ by separately fine-tuning $A$ on the forget set $\bm{\mathcal{D}_{\texttt{\bf forget}}}$ and the retain set $\bm{\mathcal{D}_{\texttt{\bf retain}}}$, respectively. Without loss of generality, we assume the base model \(A\) is significantly smaller than the reference model \(P\), making the fine-tuning process to obtain \(\smb\) and \(\smg\) substantially less resource-intensive compared to directly fine-tuning or retraining \(P\). For example, \(A\) could be a Llama2-7B model, while \(P\) could be a much larger Llama2-70B model, thus significantly reducing computational requirements.



\textbf{Unlearning via Contrastive Decoding (UCD)}. We utilize the contrastive decoding approach of \cite{li2023contrastive} to define the logits for the returned model $\mathcal{L}_{\texttt{aligned}}$, i.e. we set 
\begin{align}
\log \lma(y | x) \leftarrow \log{\lmb}(y | x) - \alpha \cdot (\log{\smb}(y | x) - \log{\smg}(y | x)) \label{eq:update}
\end{align} 
where \(\alpha > 0\) denotes a hyper-parameter (set to 0.1 in~\cite{li2023contrastive}). Correspondingly, once we have the logits corresponding to \(\lma\), we can generate next token using either: 
\begin{itemize}[leftmargin=*]
  \item \textbf{Greedy Decoding (e.g. max-sampling)}: Given an input sequence \(x\), select the next token \(y\) by choosing the one with the highest predicted probability according to model \(\lma\):
    $
    y = \arg\max_{y'} \log \lma(y' | x).
    $
     \item \textbf{Stochastic Decoding (e.g. nucleus sampling)}: Given an input sequence \(x\), randomly select the next token \(y\) based on normalized distribution given by the subset of tokens from  \(\lma\) whose cumulative probability exceeds some threshold $p$, where $p$ controls the amount of randomness.
\end{itemize}

Our unlearning update~\eqref{eq:update} modifies the logits of the reference model $\lmb$ using the difference between the logits of the auxiliary models $\smb$ and $\smg$.\footnote{While \eqref{eq:update} represents the output of contrastive decoding by $\lma$, we emphasize that no new model is computed; instead, only the logits—used to define the next-token distribution—are modified.} Recall that $\smb$ is fine-tuned on $D_{\mathrm{forget}}$, while $\smg$ is fine-tuned on $D_{\mathrm{retain}}$. The contrastive signal, defined as $\Delta_A(y \mid x) := \log \smb(y \mid x) - \log \smg(y \mid x)$, captures how much more strongly the forget-tuned model $\smb$ prefers next-token $y$ for a given prompt $x$ when compared to the retain-tuned model $\smg$. 

This contrastive signal forms the basis of our approach: we can unlearn by simply adjusting the logits of the reference model using the difference in token preferences between auxiliary models trained with and without the forget set. For illustration,  if we prompt the model with a query about a data sample that should be erased (i.e. \((x, y) \in D_{\mathrm{forget}})\), both $\Delta_A(y \mid x)$ and $\log \lmb(y \mid x)$ are likely to be high. Thus, the update in \eqref{eq:update} reduces the logit for $y$, thereby lowering its probability in the generative process and suppressing this information. More generally, when $\Delta_A(y \mid x)$ is large and positive, i.e., $\smb$ favors $y$ significantly more than $\smg$, the update decreases $\log \lmb(y \mid x)$ and thereby reduces the likelihood of generating $y$. Conversely, when $\Delta_A(y \mid x)$ is large and negative, indicating that $\smg$ prefers $y$ more than $\smb$, the update increases $\log \lmb(y \mid x)$ and thereby increases the probability of  $y$. 

\paragraph{Unlearning via Contrastive Suppression (UCS).} While UCD can both increase or decrease the probability of outputting various tokens in \(\lmb\), depending on the sign of the contrastive signal \(\Delta(y \mid x)\), in various cases, we may want to be more conservative and only make a relative decrease in logits (instead of both increasing and decreasing them using the auxiliary models). Towards that end, we also propose an update step that clips off the impact of contrastive decoding when the contrastive single is  negative: 
\begin{align*}
    \log \lma(y \mid x) \leftarrow 
\log{\lmb}(y | x) - \max\{\log{\smb}(y | x) - \log{\smg}(y | x), 0\} 
 \end{align*} 
where $\alpha > 0$ is a hyperparameter. Again, after computing the new logits, we can sample using a greedy or stochastic decoding approach.

\subsection{Why is our Contrastive Decoding Approach Effective for Unlearning? }   
We offer an initial intuition for the potential effectiveness of UCD. Although we do not present this as a comprehensive explanation of the observed behavior, we believe it sheds light on some of the underlying dynamics at play. Throughout this section, let \(\lmg\) denote the model we would have obtained (corresponding to the corrupted model \(\lmb\)) if we had trained the given reference model without the forget set \(D_{\mathrm{forget}}\).

First, as a sanity check, observe that if the auxiliary models are chosen to be the same size as the underlying models, that is, \(\smg = \lmg\) and \(\smb = \lmb\), and we set \(\alpha = 1\), then:
\begin{align}
\log \lmb(y \mid x) - \alpha \cdot \left( \log \smb(y \mid x) - \log \smg(y \mid x) \right) = \log \lmg(y \mid x).
\end{align}

In this special case, the contrastive differencing update in \eqref{eq:update} exactly recovers the next token distribution corresponding to the model  $\lmg$ that is retrained-from-scratch on the retain set.  This illustrates, in idealized conditions, how our approach enables unlearning. 

We now relax this strong equivalence assumption to examine more practical settings where the auxiliary models differ in scale or capacity from the underlying models.

\begin{prop} 
\label{prop:2}  
Suppose that for any input prompt \(x\), the auxiliary models $\smb$ and $\smg$ satisfy the relation:  \begin{align}
\log \smb(y \mid x) - \log \smg(y \mid x) \propto \log \lmb(y \mid x) -  \log \lmg(y \mid x),  \label{eq:strong_assumption} 
\end{align}
 for any token \(y \in \cY\), where  $\lmb$  denotes the initial corrupted model, and \(\lmg\) denotes the clean model (obtained by retraining-from-scratch without the  retain set). Then, there exists a choice of  \(\alpha\) that is independent of \(y\) such that the contrastive decoding  procedure in \eqref{eq:update} ensures that \(\lma \equiv  \lmg\). 
\end{prop}

The assumption in \eqref{eq:strong_assumption} formalizes the intuition that small auxiliary models can generalize the token-level preference trends observed in large models, even if the magnitude of those preferences is not preserved. Specifically, \eqref{eq:strong_assumption} suggests that if there exist tokens for which the logit difference $ \log \lmb(y \mid x) - \log \lmg(y \mid x) $ is large, indicating that the corrupted model strongly prefers token $y$ compared to the clean model (and hence $y$ should be suppressed), then a similar trend should be observable in the auxiliary models $\smb$ and $\smg$.

The proof is straightforward. Suppose the constant of proportionality in~\eqref{eq:strong_assumption} is \(m\), i.e.,
\begin{align}
\log \smb(y \mid x) - \log \smg(y \mid x) = m \left( \log \lmb(y \mid x) - \log \lmg(y \mid x) \right).
\end{align}
Then, setting \(\alpha = 1/m\) in the UCD update~\eqref{eq:update} ensures that \(\lma = \lmg\), thereby recovering the target unlearning model exactly.
While the strict proportionality in~\eqref{eq:strong_assumption} may be too strong to hold exactly in practice, the UCD update remains effective when this relationship holds approximately. Specifically, suppose there exist constants \(c_1, c_2 > 0\) such that for any token \(y\) with \(\log \lmb(y \mid x) - \log \lmg(y \mid x) \geq 0\), we have:
\begin{align}
c_1 \leq \frac{ \log \smb(y \mid x) - \log \smg(y \mid x) }{ \log \lmb(y \mid x) - \log \lmg(y \mid x) } \leq c_2. \label{eq:assumption_approximate}
\end{align}
In this case, choosing \(\alpha \in [1/c_2, 1/c_1]\) approximately aligns \(\lma\) with \(\lmg\), suppressing undesirable completions associated with the forget set while boosting completions consistent with the retain set.

\section{Experimental Setup} 

We evaluate UCD on three different tasks from two different unlearning benchmarks: Task of Fictitious Unlearning (TOFU)~\citep{maini2024tofu} and Machine Unlearning Six Ways Evaluation (MUSE)~\citep{shi2024muse}. All of our evaluations are on Llama2-13B as $\mathrm{P}$ and Llama2-7B as our auxiliary models, $\mathrm{A}$. This is the first time, to our knowledge, that existing unlearning baselines have been studied on larger models than Llama2-7B. Below we describe the specific tasks from each benchmark and the metrics used to evaluate the unlearning methods. 

\begin{figure}[t]
    \centering
    \includegraphics[width=\linewidth]{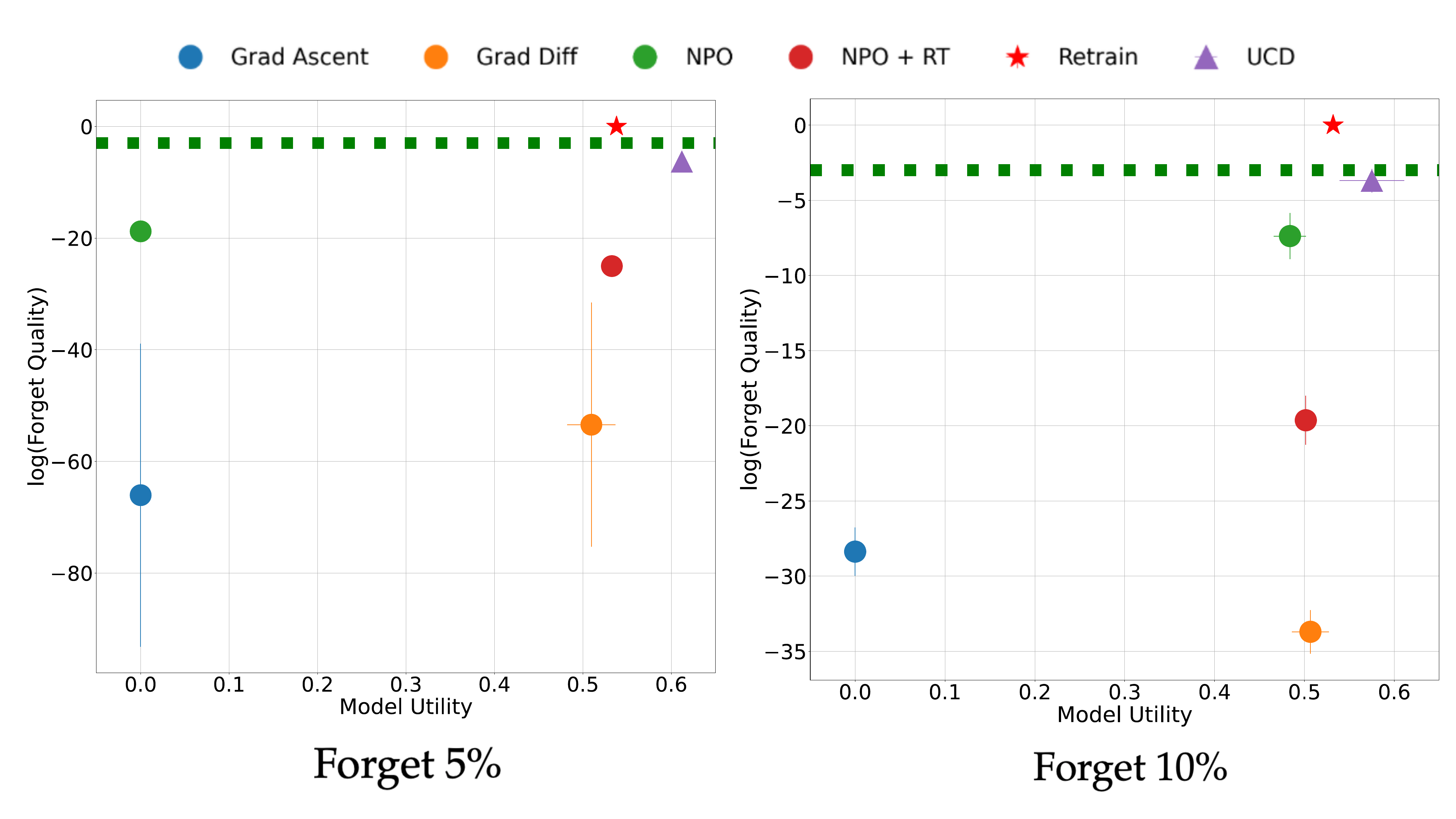}
    \caption{Forget quality versus model utility averaged over three random seeds for TOFU 5\% (left) and TOFU 10\% (right) on Llama2-13B. The dotted green line represents the forget quality ($\mathrm{log}(0.05)$) algorithms must be greater than or equal to, to be considered indistinguishable from the retrained baseline. UCD (using Llama2-7B auxiliary models) achieves the best forget quality-model utility tradeoff.}
    \label{fig:tofu_main}
\end{figure}

\paragraph{TOFU.} TOFU contains 200 GPT-4 generated author profiles, with 20 question-answer pairs for each author. The generated profiles were not contained in the pretraining data, resulting in a suitable setup for studying unlearning. We pick two tasks from TOFU: Forget 5\% and Forget 10\%, which represent forgetting 5\% and 10\% of the data, respectively. We evaluate the unlearning algorithms on these tasks according to four different sets of QA pairs: forget set, retain set, real world authors, and real world facts. More details about these sets can be found in the original TOFU paper~\citep{maini2024tofu}. We focus on measuring $\mathrm{log}$(\textit{Forget Quality}) and \textit{Model Utility} as described in~\cite{maini2024tofu}. Forget quality measures how indistinguishable the unlearned model is from the gold-standard retrained model. Indistinguishability is formalized as the $p$-value of a Kolmogorov-Smirnov test being above 0.05. Model utility measures the performance of the model on the retain set, real world authors, and real world facts sets. We report the additional metrics from TOFU of: ROUGE-L recall, probability, and truth ratio in the appendix.

\paragraph{MUSE.} MUSE represents two different corpuses of text: news articles and books. The News task contains BBC articles after 2023, and the Books task contains all of the Harry Potter books. We focus on the News task in this work because we were unable to obtain a ``clean" model for the Books task. Since we know the cutoff data for the Llama2 models this makes it easy to have clean models for the News task. Meanwhile, obtaining a clean model for Llama2 for the Books task would require pretraining a model from scratch. Similar to TOFU, we evaluate both the forget quality and model utility. For forget quality, we measure the verbatim memorization of the forget set (\texttt{VerbMem} on $\bm{\mathcal{D}_{\texttt{\bf forget}}}$), the ability to infer membership in the training data (\texttt{PrivLeak}~\citep{shi2023detecting}), and knowledge retention via QA on the forget set (\texttt{KnowMem} on $\bm{\mathcal{D}_{\texttt{\bf forget}}}$). Model utility is measured by knowledge retention via QA on the retain set (\texttt{VerbMem} on $\bm{\mathcal{D}_{\texttt{\bf retain}}})$.

\paragraph{Training and Unlearning.} For all three tasks, we compare our method against the following baselines: gradient ascent~\citep{maini2024tofu}, gradient difference~\citep{liu2022continual}, negative preference optimization (NPO)~\citep{zhang2024negative}, and NPO with a retain loss (NPO + RT). All of these baselines are described in Appendix~\ref{app:background} and were run following the open-source implementations from both the TOFU and MUSE benchmarks. We pick this subset of methods out of the ones discussed based on their performance in prior works on the chosen tasks. We average all of the results for the baselines and our method over three random seeds. We use a range of compute depending on the algorithm. Specifically, going from two NVIDIA L40s with 48GB of VRAM to 8 NVIDIA H200s with 141GB of VRAM depending on out of memory errors encountered when running on smaller amounts of compute. We elaborate more on this need to use a range of compute and how UCD is much more efficient compute wise.

\begin{table}[t]
    \centering
    \resizebox{\textwidth}{!}{%
    \begin{tabular}{lcccc}
    \toprule
    \textbf{Algorithm} & \textbf{VerbMem on} $\bm{\mathcal{D}_{\texttt{\bf forget}}} \downarrow$ & \textbf{PrivLeak} & \textbf{KnowMem on} $\bm{\mathcal{D}_{\texttt{\bf forget}}} \downarrow$ & \textbf{KnowMem on} $\bm{\mathcal{D}_{\texttt{\bf retain}}} \uparrow$  \\ 
    \midrule
    \textbf{Retrain} & {\bf \color{black} 20.99 ± 0.42 } & {\bf \color{black} 0.00 ± 0.00} &{\bf \color{black}  38.08 ± 2.13 }&{\bf \color{black}  46.15 ± 1.49 }\\
    \midrule
    \rowcolor{lightgray} UCD & {\bf 20.5 ± 0.56 } & { \bf 9.55 ± 6.65} & {\bf 36.38 ± 0.9} & {\bf 43.87 ± 1.37 }\\
    Grad Ascent & 0.0 ± 0.0 & 58.97 ± 8.25 & 0.0 ± 0.0 & 0.0 ± 0.0 \\
    \rowcolor{lightgray} Grad Diff & 0.0 ± 0.0 & -23.41 ± 3.07 & 0.0 ± 0.0 & 0.0 ± 0.0 \\
    NPO + RT & 1.02 ± 0.83 & 64.58 ± 3.22 & 28.78 ± 2.85 & 34.27 ± 2.16 \\
    \bottomrule
    \end{tabular}%
    }
    \caption{Forget quality (first three columns) versus model utility  (last column) for MUSE News. UCD achieves the best forget quality-model utility tradeoff, almost approaching the retrained model.} 
    \label{tab:muse_main}
\end{table}

\section{Results}
\subsection{UCD Improves Forgetting-Utility Tradeoff}
\label{res:main}

UCD significantly outperforms the baselines across all three tasks. We report the best performing UCD model based on tuning of $\alpha$ over $\{0.01, 0.1, 0.5, 1.0\}$. As shown in Figure~\ref{fig:tofu_main}, for TOFU 5\%, UCD achieves indistinguishability from the retrained model and also improves the model utility. We believe that the improvement in utility compared to the retrained model can be attributed to the contrastive decoding approach. Numerous prior works show contrastive decoding improves text quality and diversity~\citep{li2023contrastive, obrien2023contrastive}. We provide an example of how UCD successfully recovers the retrained model compared to all other baselines in Appendix~\ref{app:example_gen}. For MUSE, UCD is the closest model to replicating the retrained model (Table~\ref{tab:muse_main}). UCD overcomes issues of over unlearning / under unlearning (measured by the \texttt{PrivLeak} metric) and poor model utility discussed in the original paper.

\subsection{Bootsrapping from Existing Unlearning Algorithms Improves Tradeoff}

Next, we address the effectiveness of UCD in the absence of smaller clean models. In this setting, we approximate the clean model—i.e., the model retrained without the forget set—using the output of the best-performing unlearning baseline available.  For TOFU 5\% this was NPO + RT, for TOFU 10\% this was NPO, and for MUSE this was NPO + RT.

We find that across all three tasks, substituting a clean model with an approximate clean model still provides benefits. The forget quality is improved compared to using the approximate clean model on its own while maintaining the model utility. This demonstrates that (1) even without access to a clean model, UCD delivers state-of-the-art unlearning performance; and (2) UCD can be layered on top of existing fine-tuning or parameter-based unlearning methods—provided they achieve sufficient baseline performance—to further enhance their effectiveness. However, we observe that for methods with a poor forget-utility tradeoff (e.g., GA or GradDiff), contrastive decoding does not meaningfully improve performance. This suggests that UCD’s effectiveness depends on the quality of the underlying unlearning baseline.

\begin{figure}[h]
    \centering
    \begin{minipage}[b]{0.48\textwidth}
        \centering
        \includegraphics[width=\textwidth]{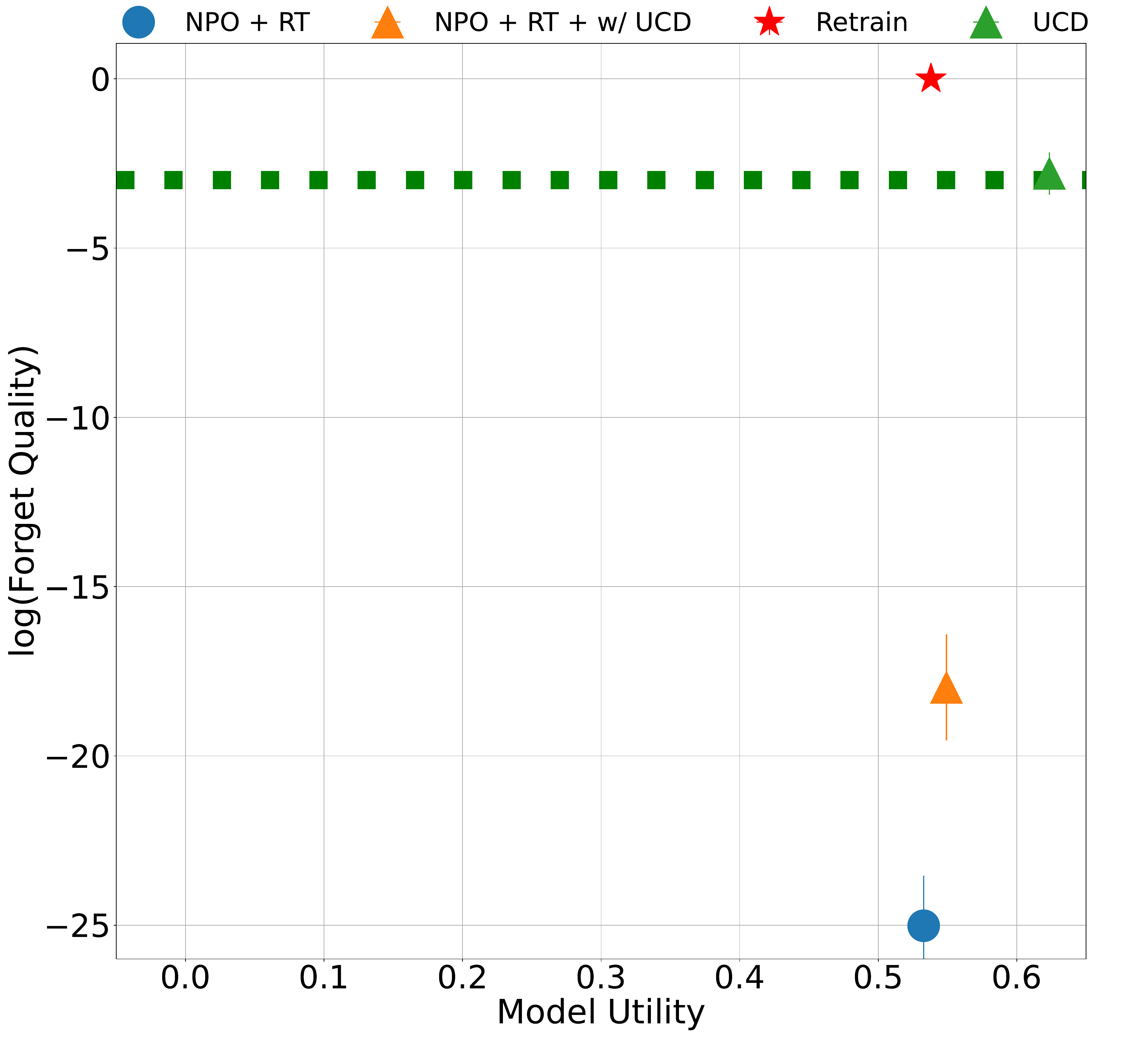}
        \caption{Forget 5\%}
        \label{fig:plot1}
    \end{minipage}
    \hfill
    \begin{minipage}[b]{0.48\textwidth}
        \centering
        \includegraphics[width=\textwidth]{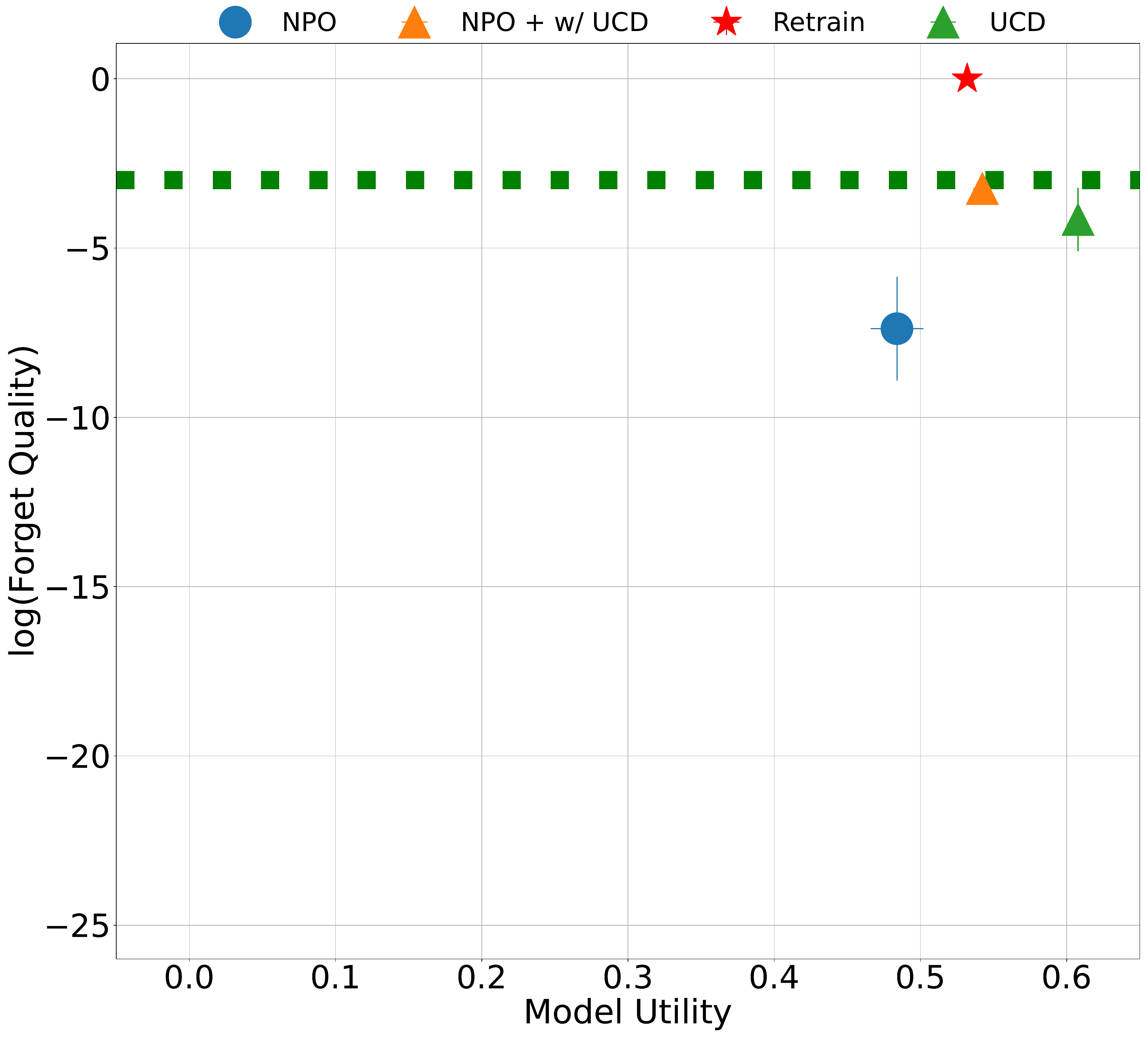}
        \caption{Forget 10\%}
        \label{fig:plot2}
    \end{minipage}
    \caption{Forget quality versus model utility for TOFU 5\% (left) and TOFU 10\% (right) on Llama2-13B when using the best performing approximate clean models (i.e. NPO + RT (left) and NPO (right)) instead of exact clean models. Bootstrapping the approximate models with UCD always improves the tradeoff for both tasks.}\label{fig:tofu_bootstrap}
\end{figure}

\subsection{ UCD \& UCS Scale to Very Large Models}
\label{sec:scale}

\begin{figure}[t]
    \centering
    \includegraphics[width=0.8\linewidth]{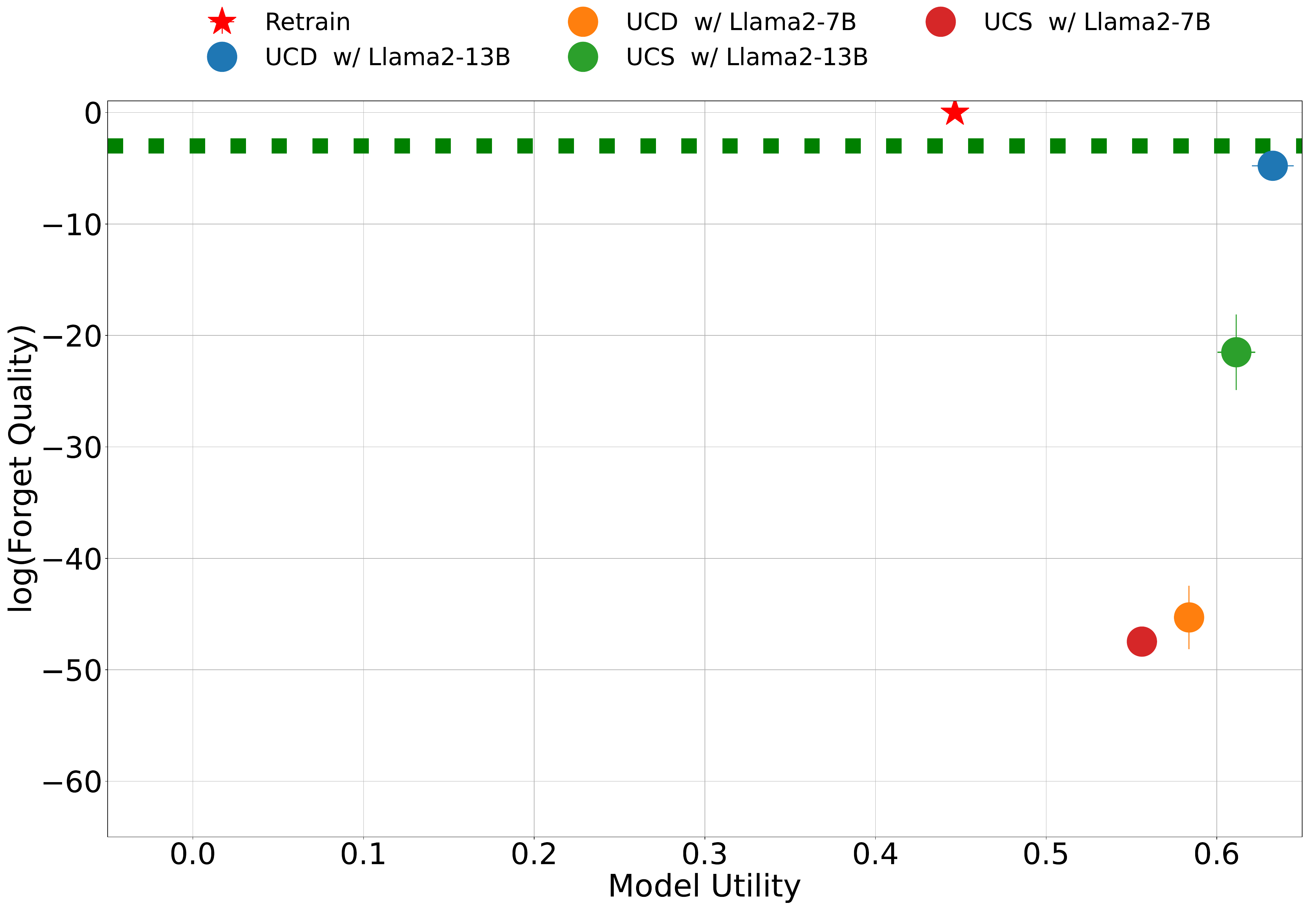}
    \caption{Forget quality versus model utility TOFU 10\% on Llama2-70B. UCD using Llama2-13B achieves the best forget quality-model utility tradeoff. Even improving upon the model utility of the retrained model.}
    \label{fig:tofu_scale}
\end{figure}

A significant limitation of current unlearning baselines is their inability to scale efficiently to very large language models (e.g. beyond 7B and 13B) without extensive compute resources. Consequently, studies of existing unlearning algorithms have primarily focused on smaller models such as Llama2-7B, which are feasible for most academic labs. Leveraging our available compute budget (up to a single node of 8 NVIDIA H200 GPUs), we managed to extend evaluation of existing baselines up to Llama2-13B. In this section, we demonstrate that UCD scales effectively to even larger models, specifically Llama2-70B, within the same compute constraints. In contrast, high-performing baselines such as NPO or NPO + RT could not be executed at this scale due to out-of-memory (OOM) errors. As illustrated in Figure~\ref{fig:tofu_scale}, when employing Llama2-13B as auxiliary models, UCD closely approximates the forget performance of the retrained model and notably enhances utility (from approximately 45\% to 62\%) compared to retraining. Furthermore, as shown in Table~\ref{tab:tofu_scale_compute}, UCD offers optimal training and inference efficiency, enabling practical scaling to Llama2-70B models.

\section{Ablations}

\vspace{-0.5em}
\subsection{Sensitivity to Sampling Strategy and  Hyperparameter-\(\alpha\)} 

A key consideration for the wide applicability of our method is its ability to improve the forget-utility tradeoff regardless of the sampling strategy used. We examine two commonly used sampling strategies in production LLMs: greedy and top-\(p\) (nucleus) sampling~\citep{holtzman2019curious}, where \(p\) is set to either 0.7 or 0.9. Recreating the plots and tables from Section~\ref{res:main} with each sampling strategy, we  find that UCD outperforms most methods. For TOFU, since many of the metrics are computed using the loss, the results are identical between greedy and top-\(p\) sampling. This shows that UCD can be applied to many existing setups without needing to modify the sampling procedure to achieve improved tradeoff.

We also investigate the sensitivity of UCD to the alpha parameter. We find that for TOFU ideal values are either 0.5 or 1.0 depending on the task. Values lower than 0.5 tended to be too low and resulted in poor forget quality (Figure~\ref{fig:tofu_alpha}). For MUSE, an alpha value of 1.0 yielded the best performance (Table~\ref{tab:muse_alpha}).

\vspace{-0.7em}
\subsection{Suppression vs. Differencing} 

Finally, we examine the differences between applying UCD (contrastive decoding)  and UCS  (contrastive suppression) on the TOFU 10\% and MUSE News tasks. We already demonstrated some of this difference in Section~\ref{sec:scale}. Contrastive decoding achieves the strongest forget-quality versus model-utility tradeoff when clean auxiliary models are available~(Figure~\ref{fig:tofu_compare}). However, in scenarios where bootstrapping is necessary, i.e., when clean models are replaced by approximations, contrastive suppression tends to yield better tradeoffs, effectively improving forget quality without negatively impacting utility (Table~\ref{tab:muse_diff_vs_max}). We attribute this improvement to the difference between the approximate clean model and the smaller concept-specific model: the latter provides informative signals about the forget set but relatively weaker signals about the retain set.

\ascomment{Table 1 has arrows in the first row, but Table 2 does not. Also, fix the caption for Table 2.}

\section{Discussion and Conclusion} 
Our proposed method, UCD, offers several significant advantages over existing unlearning approaches. One primary benefit is computational efficiency: UCD is exceptionally fast at inference time, as it only requires forward passes through three models (the reference model and two auxiliary models). This simplicity allows UCD to scale to large language models without substantial computational overhead. Additionally, UCD relies solely on a gray-box update mechanism, requiring access only to logits from the relevant models ($\lmb$, $\smb$, and $\smg$), rather than requiring full access to their parameters or gradients.

Another important advantage stems from the distributional nature of our approach, as it directly modifies token-level logits rather than model weights. Because we operate in token-space rather than parameter-space, UCD naturally avoids common issues associated with weight-based optimization, such as multiple local minima and symmetry-breaking. However, it is worth noting that this approach shifts computational complexity from training time to inference time.

Despite these benefits, UCD also faces several important limitations. Most notably, our approach currently lacks rigorous theoretical guarantees beyond the simplistic setting of Proposition \ref{prop:2}, as well as a formal definition of unlearning suitable for generative language models. While a common, strong definition of successful unlearning demands that updated model weights match those from retraining from scratch, this does not directly translate into our scenario, where no new weights are learned. Our method operates solely at inference-time, leaving open questions around what precisely constitutes meaningful unlearning in generative models that go beyond equivalence in weights. 

Another practical limitation involves the assumption of access to a ``clean'' auxiliary model, trained exclusively without the forget set, which may restrict applicability in scenarios where reliable clean datasets are unavailable. Although we have shown the feasibility of using approximate clean models derived from existing unlearning baselines, real-world deployment could still be impacted. Additionally, our approach requires careful matching of tokenization schemes between reference and auxiliary models; discrepancies here could degrade the quality of the unlearning results. 


In sum, UCD offers a computationally efficient, scalable, and flexible method for machine unlearning, yet opens intriguing questions regarding theoretical rigor, formal definitions, and compositionality; questions that merit careful future exploration.

\ascomment{Add a discussion on distillation!}

 \section*{Ethics Statement}
Like all unlearning techniques, UCD relies on auxiliary models trained on partitioned datasets. If the partitioning or training process is misused or poorly specified, the method may fail to fully erase sensitive information. Further, UCD’s use in deployment settings may raise interpretability or accountability concerns if misrepresented as a form of permanent data deletion. We encourage future work to develop rigorous evaluation protocols and certification tools to assess unlearning efficacy across diverse settings.
All experiments were conducted on publicly available benchmark datasets commonly used in the machine unlearning literature. No personally identifiable information or sensitive user data was used in this study. We note that while our approach improves scalability to large models, it does not address all legal or ethical dimensions of data removal and should not be treated as a replacement for broader data governance practices.

\section*{Acknowledgements} 
We thank Angelos Assos and Zhili Feng for the useful discussions. ACW acknowledges support from Simons Collaboration on Algorithmic Fairness and MIT Generative AI Impact Award. AS acknowledges support from ARO through award W911NF-21-1-0328, as well as the Simons Foundation
and the NSF through award DMS-2031883. 

\bibliography{colm2025_conference}

\begin{thebibliography}{58}
\providecommand{\natexlab}[1]{#1}
\providecommand{\url}[1]{\texttt{#1}}
\expandafter\ifx\csname urlstyle\endcsname\relax
  \providecommand{\doi}[1]{doi: #1}\else
  \providecommand{\doi}{doi: \begingroup \urlstyle{rm}\Url}\fi

\bibitem[CCP(2018)]{CCPA2018}
California consumer privacy act of 2018.
\newblock \url{https://leginfo.legislature.ca.gov/faces/billTextClient.xhtml?bill_id=201720180AB375}, 2018.

\bibitem[Barez et~al.(2025)Barez, Fu, Prabhu, Casper, Sanyal, Bibi, O'Gara, Kirk, Bucknall, Fist, et~al.]{barez2025open}
Fazl Barez, Tingchen Fu, Ameya Prabhu, Stephen Casper, Amartya Sanyal, Adel Bibi, Aidan O'Gara, Robert Kirk, Ben Bucknall, Tim Fist, et~al.
\newblock Open problems in machine unlearning for ai safety.
\newblock \emph{arXiv preprint arXiv:2501.04952}, 2025.

\bibitem[Bourtoule et~al.(2021)Bourtoule, Chandrasekaran, Choquette-Choo, Jia, Travers, Zhang, Lie, and Papernot]{bourtoule2021machine}
Lucas Bourtoule, Varun Chandrasekaran, Christopher~A Choquette-Choo, Hengrui Jia, Adelin Travers, Baiwu Zhang, David Lie, and Nicolas Papernot.
\newblock Machine unlearning.
\newblock In \emph{2021 IEEE Symposium on Security and Privacy (SP)}, pp.\  141--159. IEEE, 2021.

\bibitem[Cao \& Yang(2015)Cao and Yang]{cao2015towards}
Yinzhi Cao and Junfeng Yang.
\newblock Towards making systems forget with machine unlearning.
\newblock In \emph{2015 IEEE symposium on security and privacy}, pp.\  463--480. IEEE, 2015.

\bibitem[Carlini et~al.(2021)Carlini, Tramer, Wallace, Jagielski, Herbert-Voss, Lee, Roberts, Brown, Song, Erlingsson, et~al.]{carlini2021extracting}
Nicholas Carlini, Florian Tramer, Eric Wallace, Matthew Jagielski, Ariel Herbert-Voss, Katherine Lee, Adam Roberts, Tom Brown, Dawn Song, Ulfar Erlingsson, et~al.
\newblock Extracting training data from large language models.
\newblock In \emph{30th USENIX security symposium (USENIX Security 21)}, pp.\  2633--2650, 2021.

\bibitem[Chen \& Yang(2023)Chen and Yang]{chen2023unlearn}
Jiaao Chen and Diyi Yang.
\newblock Unlearn what you want to forget: Efficient unlearning for llms.
\newblock \emph{arXiv preprint arXiv:2310.20150}, 2023.

\bibitem[Chowdhury et~al.(2024)Chowdhury, Choromanski, Sehanobish, Dubey, and Chaturvedi]{chowdhury2024towards}
Somnath Basu~Roy Chowdhury, Krzysztof Choromanski, Arijit Sehanobish, Avinava Dubey, and Snigdha Chaturvedi.
\newblock Towards scalable exact machine unlearning using parameter-efficient fine-tuning.
\newblock \emph{arXiv preprint arXiv:2406.16257}, 2024.

\bibitem[Eldan \& Russinovich(2023)Eldan and Russinovich]{eldan2023whos}
Ronen Eldan and Mark Russinovich.
\newblock Who's harry potter? approximate unlearning in llms, 2023.

\bibitem[{European Union}(2016)]{gdpr2016}
{European Union}.
\newblock Regulation (eu) 2016/679 of the european parliament and of the council of 27 april 2016.
\newblock \url{https://gdpr-info.eu}, 2016.
\newblock General Data Protection Regulation (GDPR).

\bibitem[Fan et~al.()Fan, Liu, Lin, Jia, Zhang, Mei, and Liu]{fan2024simplicity}
Chongyu Fan, Jiancheng Liu, Licong Lin, Jinghan Jia, Ruiqi Zhang, Song Mei, and Sijia Liu.
\newblock Simplicity prevails: Rethinking negative preference optimization for llm unlearning.
\newblock In \emph{Neurips Safe Generative AI Workshop 2024}.

\bibitem[Ghazi et~al.(2023)Ghazi, Kamath, Kumar, Manurangsi, Sekhari, and Zhang]{ghazi2023ticketed}
Badih Ghazi, Pritish Kamath, Ravi Kumar, Pasin Manurangsi, Ayush Sekhari, and Chiyuan Zhang.
\newblock Ticketed learning--unlearning schemes.
\newblock In \emph{The Thirty Sixth Annual Conference on Learning Theory}, pp.\  5110--5139. PMLR, 2023.

\bibitem[Gupta et~al.(2021)Gupta, Jung, Neel, Roth, Sharifi-Malvajerdi, and Waites]{gupta2021adaptive}
Varun Gupta, Christopher Jung, Seth Neel, Aaron Roth, Saeed Sharifi-Malvajerdi, and Chris Waites.
\newblock Adaptive machine unlearning.
\newblock \emph{Advances in Neural Information Processing Systems}, 34:\penalty0 16319--16330, 2021.

\bibitem[Holtzman et~al.(2019)Holtzman, Buys, Du, Forbes, and Choi]{holtzman2019curious}
Ari Holtzman, Jan Buys, Li~Du, Maxwell Forbes, and Yejin Choi.
\newblock The curious case of neural text degeneration.
\newblock \emph{arXiv preprint arXiv:1904.09751}, 2019.

\bibitem[Ilharco et~al.(2022)Ilharco, Ribeiro, Wortsman, Gururangan, Schmidt, Hajishirzi, and Farhadi]{ilharco2022editing}
Gabriel Ilharco, Marco~Tulio Ribeiro, Mitchell Wortsman, Suchin Gururangan, Ludwig Schmidt, Hannaneh Hajishirzi, and Ali Farhadi.
\newblock Editing models with task arithmetic.
\newblock \emph{arXiv preprint arXiv:2212.04089}, 2022.

\bibitem[Ishibashi \& Shimodaira(2023)Ishibashi and Shimodaira]{ishibashi2023knowledge}
Yoichi Ishibashi and Hidetoshi Shimodaira.
\newblock Knowledge sanitization of large language models.
\newblock \emph{arXiv preprint arXiv:2309.11852}, 2023.

\bibitem[Jang et~al.(2022)Jang, Yoon, Yang, Cha, Lee, Logeswaran, and Seo]{jang2022knowledge}
Joel Jang, Dongkeun Yoon, Sohee Yang, Sungmin Cha, Moontae Lee, Lajanugen Logeswaran, and Minjoon Seo.
\newblock Knowledge unlearning for mitigating privacy risks in language models.
\newblock \emph{arXiv preprint arXiv:2210.01504}, 2022.

\bibitem[Jia et~al.(2024)Jia, Zhang, Zhang, Liu, Runwal, Diffenderfer, Kailkhura, and Liu]{jia2024soul}
Jinghan Jia, Yihua Zhang, Yimeng Zhang, Jiancheng Liu, Bharat Runwal, James Diffenderfer, Bhavya Kailkhura, and Sijia Liu.
\newblock Soul: Unlocking the power of second-order optimization for llm unlearning.
\newblock \emph{arXiv preprint arXiv:2404.18239}, 2024.

\bibitem[Kadhe et~al.(2024)Kadhe, Ahmed, Wei, Baracaldo, and Padhi]{kadhe2024split}
Swanand~Ravindra Kadhe, Farhan Ahmed, Dennis Wei, Nathalie Baracaldo, and Inkit Padhi.
\newblock Split, unlearn, merge: Leveraging data attributes for more effective unlearning in llms.
\newblock \emph{arXiv preprint arXiv:2406.11780}, 2024.

\bibitem[Karamolegkou et~al.(2023)Karamolegkou, Li, Zhou, and S{\o}gaard]{karamolegkou2023copyright}
Antonia Karamolegkou, Jiaang Li, Li~Zhou, and Anders S{\o}gaard.
\newblock Copyright violations and large language models.
\newblock \emph{arXiv preprint arXiv:2310.13771}, 2023.

\bibitem[Karp(2023)]{karp2023australia}
Pual Karp.
\newblock Australia to consider european-style right to be forgotten privacy laws.
\newblock \emph{The Guardian (Jan. 19, 2023).(Visited on 01/19/2023)}, 2023.

\bibitem[Kumar et~al.(2022)Kumar, Gangadharaiah, and Roth]{kumar2022privacy}
Vinayshekhar~Bannihatti Kumar, Rashmi Gangadharaiah, and Dan Roth.
\newblock Privacy adhering machine un-learning in nlp.
\newblock \emph{arXiv preprint arXiv:2212.09573}, 2022.

\bibitem[Kuo et~al.()Kuo, Setlur, Srinivas, Raghunathan, and Smith]{kuo2025exact}
Kevin Kuo, Amrith Setlur, Kartik Srinivas, Aditi Raghunathan, and Virginia Smith.
\newblock Exact unlearning of finetuning data via model merging at scale.
\newblock In \emph{ICLR 2025 Workshop on Modularity for Collaborative, Decentralized, and Continual Deep Learning}.

\bibitem[Kurmanji et~al.(2023)Kurmanji, Triantafillou, Hayes, and Triantafillou]{kurmanji2023towards}
Meghdad Kurmanji, Peter Triantafillou, Jamie Hayes, and Eleni Triantafillou.
\newblock Towards unbounded machine unlearning.
\newblock \emph{Advances in neural information processing systems}, 36:\penalty0 1957--1987, 2023.

\bibitem[Lev \& Wilson(2024)Lev and Wilson]{lev2024faster}
Omri Lev and Ashia Wilson.
\newblock Faster machine unlearning via natural gradient descent.
\newblock \emph{arXiv preprint arXiv:2407.08169}, 2024.

\bibitem[Li et~al.(2024)Li, Pan, Gopal, Yue, Berrios, Gatti, Li, Dombrowski, Goel, Phan, et~al.]{li2024wmdp}
Nathaniel Li, Alexander Pan, Anjali Gopal, Summer Yue, Daniel Berrios, Alice Gatti, Justin~D Li, Ann-Kathrin Dombrowski, Shashwat Goel, Long Phan, et~al.
\newblock The wmdp benchmark: Measuring and reducing malicious use with unlearning.
\newblock \emph{arXiv preprint arXiv:2403.03218}, 2024.

\bibitem[Li et~al.(2023)Li, Holtzman, Fried, Liang, Eisner, Hashimoto, Zettlemoyer, and Lewis]{li2023contrastive}
Xiang~Lisa Li, Ari Holtzman, Daniel Fried, Percy Liang, Jason Eisner, Tatsunori Hashimoto, Luke Zettlemoyer, and Mike Lewis.
\newblock Contrastive decoding: Open-ended text generation as optimization.
\newblock In \emph{Proceedings of the 61st Annual Meeting of the Association for Computational Linguistics (ACL)}, pp.\  12286--12312. Association for Computational Linguistics, 2023.
\newblock \doi{10.18653/v1/2023.acl-long.687}.
\newblock URL \url{https://aclanthology.org/2023.acl-long.687/}.

\bibitem[Liu et~al.(2022)Liu, Liu, and Stone]{liu2022continual}
Bo~Liu, Qiang Liu, and Peter Stone.
\newblock Continual learning and private unlearning.
\newblock In \emph{Conference on Lifelong Learning Agents}, pp.\  243--254. PMLR, 2022.

\bibitem[Liu et~al.(2024{\natexlab{a}})Liu, Wang, Flanigan, and Liu]{liu2024large}
Chris~Yuhao Liu, Yaxuan Wang, Jeffrey Flanigan, and Yang Liu.
\newblock Large language model unlearning via embedding-corrupted prompts.
\newblock \emph{arXiv preprint arXiv:2406.07933}, 2024{\natexlab{a}}.

\bibitem[Liu et~al.(2024{\natexlab{b}})Liu, Yao, Jia, Casper, Baracaldo, Hase, Xu, Yao, Li, Varshney, et~al.]{liu2024rethinking}
Sijia Liu, Yuanshun Yao, Jinghan Jia, Stephen Casper, Nathalie Baracaldo, Peter Hase, Xiaojun Xu, Yuguang Yao, Hang Li, Kush~R Varshney, et~al.
\newblock Rethinking machine unlearning for large language models.
\newblock \emph{arXiv preprint arXiv:2402.08787}, 2024{\natexlab{b}}.

\bibitem[Lu et~al.(2022)Lu, Welleck, Hessel, Jiang, Qin, West, Ammanabrolu, and Choi]{lu2022quark}
Ximing Lu, Sean Welleck, Jack Hessel, Liwei Jiang, Lianhui Qin, Peter West, Prithviraj Ammanabrolu, and Yejin Choi.
\newblock Quark: Controllable text generation with reinforced unlearning.
\newblock \emph{Advances in neural information processing systems}, 35:\penalty0 27591--27609, 2022.

\bibitem[{\L}ucki et~al.(2024){\L}ucki, Wei, Huang, Henderson, Tram{\`e}r, and Rando]{lucki2024adversarial}
Jakub {\L}ucki, Boyi Wei, Yangsibo Huang, Peter Henderson, Florian Tram{\`e}r, and Javier Rando.
\newblock An adversarial perspective on machine unlearning for ai safety.
\newblock \emph{arXiv preprint arXiv:2409.18025}, 2024.

\bibitem[Maini et~al.(2024)Maini, Feng, Schwarzschild, Lipton, and Kolter]{maini2024tofu}
Pratyush Maini, Zhili Feng, Avi Schwarzschild, Zachary~C. Lipton, and J.~Zico Kolter.
\newblock Tofu: A task of fictitious unlearning for llms, 2024.

\bibitem[Meng et~al.(2022{\natexlab{a}})Meng, Bau, Andonian, and Belinkov]{meng2022rome}
Kevin Meng, David Bau, Alex Andonian, and Yonatan Belinkov.
\newblock Locating and editing factual associations in gpt.
\newblock \emph{arXiv preprint arXiv:2202.05262}, 2022{\natexlab{a}}.

\bibitem[Meng et~al.(2022{\natexlab{b}})Meng, Sharma, Andonian, Belinkov, and Bau]{meng2022mass}
Kevin Meng, Arnab~Sen Sharma, Alex Andonian, Yonatan Belinkov, and David Bau.
\newblock Mass-editing memory in a transformer.
\newblock \emph{arXiv preprint arXiv:2210.07229}, 2022{\natexlab{b}}.

\bibitem[Meng et~al.(2023)Meng, Bau, Andonian, and Belinkov]{meng2023memit}
Kevin Meng, David Bau, Alex Andonian, and Yonatan Belinkov.
\newblock Mass editing memory in a transformer.
\newblock \emph{arXiv preprint arXiv:2302.09232}, 2023.

\bibitem[Mitchell et~al.(2022{\natexlab{a}})Mitchell, Lin, Bosselut, Finn, and Manning]{mitchell2022serac}
Eric Mitchell, Charles Lin, Antoine Bosselut, Chelsea Finn, and Christopher~D. Manning.
\newblock Memory-based model editing at scale.
\newblock \emph{arXiv preprint arXiv:2110.11309}, 2022{\natexlab{a}}.

\bibitem[Mitchell et~al.(2022{\natexlab{b}})Mitchell, Lin, Bosselut, Manning, and Finn]{mitchell2022memory}
Eric Mitchell, Charles Lin, Antoine Bosselut, Christopher~D Manning, and Chelsea Finn.
\newblock Memory-based model editing at scale.
\newblock In \emph{International Conference on Machine Learning}, pp.\  15817--15831. PMLR, 2022{\natexlab{b}}.

\bibitem[Neel et~al.(2021)Neel, Roth, and Sharifi-Malvajerdi]{neel2021descent}
Seth Neel, Aaron Roth, and Saeed Sharifi-Malvajerdi.
\newblock Descent-to-delete: Gradient-based methods for machine unlearning.
\newblock In \emph{Algorithmic Learning Theory}, pp.\  931--962. PMLR, 2021.

\bibitem[Nguyen et~al.(2022)Nguyen, Huynh, Nguyen, Liew, Yin, and Nguyen]{nguyen2022survey}
Thanh~Tam Nguyen, Thanh~Trung Huynh, Phi~Le Nguyen, Alan Wee-Chung Liew, Hongzhi Yin, and Quoc Viet~Hung Nguyen.
\newblock A survey of machine unlearning.
\newblock \emph{arXiv preprint arXiv:2209.02299}, 2022.

\bibitem[O'Brien \& Lewis(2023)O'Brien and Lewis]{obrien2023contrastive}
Sean O'Brien and Mike Lewis.
\newblock Contrastive decoding improves reasoning in large language models.
\newblock \emph{arXiv preprint arXiv:2309.09117}, 2023.
\newblock URL \url{https://arxiv.org/abs/2309.09117}.

\bibitem[Pawelczyk et~al.(2023)Pawelczyk, Neel, and Lakkaraju]{pawelczyk2023context}
Martin Pawelczyk, Seth Neel, and Himabindu Lakkaraju.
\newblock In-context unlearning: Language models as few shot unlearners.
\newblock \emph{arXiv preprint arXiv:2310.07579}, 2023.

\bibitem[Rafailov et~al.(2024)Rafailov, Sharma, Mitchell, Manning, Ermon, and Finn]{rafailov2024direct}
Rafael Rafailov, Archit Sharma, Eric Mitchell, Christopher~D Manning, Stefano Ermon, and Chelsea Finn.
\newblock Direct preference optimization: Your language model is secretly a reward model.
\newblock \emph{Advances in Neural Information Processing Systems}, 36, 2024.

\bibitem[Schwarzschild et~al.(2024)Schwarzschild, Feng, Maini, Lipton, and Kolter]{schwarzschild2024rethinking}
Avi Schwarzschild, Zhili Feng, Pratyush Maini, Zachary~C Lipton, and J~Zico Kolter.
\newblock Rethinking llm memorization through the lens of adversarial compression.
\newblock \emph{arXiv preprint arXiv:2404.15146}, 2024.

\bibitem[Sekhari et~al.(2021)Sekhari, Acharya, Kamath, and Suresh]{sekhari2021remember}
Ayush Sekhari, Jayadev Acharya, Gautam Kamath, and Ananda~Theertha Suresh.
\newblock Remember what you want to forget: Algorithms for machine unlearning.
\newblock \emph{Advances in Neural Information Processing Systems}, 34:\penalty0 18075--18086, 2021.

\bibitem[Shevlane et~al.(2023)Shevlane, Farquhar, Garfinkel, Phuong, Whittlestone, Leung, Kokotajlo, Marchal, Anderljung, Kolt, et~al.]{shevlane2023model}
Toby Shevlane, Sebastian Farquhar, Ben Garfinkel, Mary Phuong, Jess Whittlestone, Jade Leung, Daniel Kokotajlo, Nahema Marchal, Markus Anderljung, Noam Kolt, et~al.
\newblock Model evaluation for extreme risks.
\newblock \emph{arXiv preprint arXiv:2305.15324}, 2023.

\bibitem[Shi et~al.(2023)Shi, Ajith, Xia, Huang, Liu, Blevins, Chen, and Zettlemoyer]{shi2023detecting}
Weijia Shi, Anirudh Ajith, Mengzhou Xia, Yangsibo Huang, Daogao Liu, Terra Blevins, Danqi Chen, and Luke Zettlemoyer.
\newblock Detecting pretraining data from large language models.
\newblock \emph{arXiv preprint arXiv:2310.16789}, 2023.

\bibitem[Shi et~al.(2024)Shi, Lee, Huang, Malladi, Zhao, Holtzman, Liu, Zettlemoyer, Smith, and Zhang]{shi2024muse}
Weijia Shi, Jaechan Lee, Yangsibo Huang, Sadhika Malladi, Jieyu Zhao, Ari Holtzman, Daogao Liu, Luke Zettlemoyer, Noah~A Smith, and Chiyuan Zhang.
\newblock Muse: Machine unlearning six-way evaluation for language models.
\newblock \emph{arXiv preprint arXiv:2407.06460}, 2024.

\bibitem[Suriyakumar \& Wilson(2022)Suriyakumar and Wilson]{suriyakumar2022algorithms}
Vinith Suriyakumar and Ashia~C Wilson.
\newblock Algorithms that approximate data removal: New results and limitations.
\newblock \emph{Advances in Neural Information Processing Systems}, 35:\penalty0 18892--18903, 2022.

\bibitem[Thaker et~al.(2024)Thaker, Maurya, and Smith]{thaker2024guardrail}
Pratiksha Thaker, Yash Maurya, and Virginia Smith.
\newblock Guardrail baselines for unlearning in llms.
\newblock \emph{arXiv preprint arXiv:2403.03329}, 2024.

\bibitem[Voigt \& Von~dem Bussche(2017)Voigt and Von~dem Bussche]{voigt2017eu}
Paul Voigt and Axel Von~dem Bussche.
\newblock The eu general data protection regulation (gdpr).
\newblock \emph{A practical guide, 1st ed., Cham: Springer International Publishing}, 10\penalty0 (3152676):\penalty0 10--5555, 2017.

\bibitem[Wang et~al.(2024)Wang, Wu, He, Chen, and McAuley]{wang2024large}
Yu~Wang, Ruihan Wu, Zexue He, Xiusi Chen, and Julian McAuley.
\newblock Large scale knowledge washing.
\newblock \emph{arXiv preprint arXiv:2405.16720}, 2024.

\bibitem[Xu et~al.(2024)Xu, Sehwag, Koppel, Zhu, An, Huang, and Ganesh]{xu2024genarm}
Yuancheng Xu, Udari~Madhushani Sehwag, Alec Koppel, Sicheng Zhu, Bang An, Furong Huang, and Sumitra Ganesh.
\newblock Genarm: Reward guided generation with autoregressive reward model for test-time alignment.
\newblock \emph{arXiv preprint arXiv:2410.08193}, 2024.

\bibitem[Yang et~al.(2024)Yang, Shen, Guo, Wang, Cao, Zhang, and Tao]{yang2024model}
Enneng Yang, Li~Shen, Guibing Guo, Xingwei Wang, Xiaochun Cao, Jie Zhang, and Dacheng Tao.
\newblock Model merging in llms, mllms, and beyond: Methods, theories, applications and opportunities.
\newblock \emph{arXiv preprint arXiv:2408.07666}, 2024.

\bibitem[Yao et~al.(2023{\natexlab{a}})Yao, Xu, and Liu]{yao2023large}
Yuanshun Yao, Xiaojun Xu, and Yang Liu.
\newblock Large language model unlearning.
\newblock \emph{arXiv preprint arXiv:2310.10683}, 2023{\natexlab{a}}.

\bibitem[Yao et~al.(2023{\natexlab{b}})Yao, Wang, Tian, Cheng, Li, Deng, Chen, and Zhang]{yao2023editing}
Yunzhi Yao, Peng Wang, Bozhong Tian, Siyuan Cheng, Zhoubo Li, Shumin Deng, Huajun Chen, and Ningyu Zhang.
\newblock Editing large language models: Problems, methods, and opportunities.
\newblock \emph{arXiv preprint arXiv:2305.13172}, 2023{\natexlab{b}}.

\bibitem[Zhang et~al.(2023{\natexlab{a}})Zhang, Zhang, Zhang, and Kweon]{zhang2023text}
Chenshuang Zhang, Chaoning Zhang, Mengchun Zhang, and In~So Kweon.
\newblock Text-to-image diffusion models in generative ai: A survey.
\newblock \emph{arXiv preprint arXiv:2303.07909}, 2023{\natexlab{a}}.

\bibitem[Zhang et~al.(2023{\natexlab{b}})Zhang, Wang, Xu, Wang, and Shi]{zhang2023forget}
Eric Zhang, Kai Wang, Xingqian Xu, Zhangyang Wang, and Humphrey Shi.
\newblock Forget-me-not: Learning to forget in text-to-image diffusion models.
\newblock \emph{arXiv preprint arXiv:2303.17591}, 2023{\natexlab{b}}.

\bibitem[Zhang et~al.(2024)Zhang, Lin, Bai, and Mei]{zhang2024negative}
Ruiqi Zhang, Licong Lin, Yu~Bai, and Song Mei.
\newblock Negative preference optimization: From catastrophic collapse to effective unlearning.
\newblock \emph{arXiv preprint arXiv:2404.05868}, 2024.

\end{thebibliography}
\bibliographystyle{colm2025_conference}

\appendix

\clearpage 

\section{Background and Additional Related Work} \label{app:background}
\ascomment{For Arxiv: Move this to the main body!} 

Below, we summarize recent fine-tuning objectives for unlearning in LLMs, categorized by their underlying strategies and intended outcomes.

\begin{itemize}[left=0pt]
\item \textbf{Gradient Ascent (GA)}: A common unlearning baseline that maximizes the next-token prediction loss on the forget set to reverse learning on those examples:
    \begin{equation*}
        \mathcal{L}_{\texttt{GA}}(\theta) =  \mathbb{E}_{x\sim\bm{\mathcal{D}_{\texttt{\bf forget}}}}[\log(\pi_\theta(y|x))].
    \end{equation*}
    While simple and direct, GA often serves as the foundation for more stable and effective variants:

    \item \textbf{Gradient Difference (GD)}: Extends GA by adding a standard training loss on the retain set to preserve performance:
    \begin{equation*}
        \mathcal{L}_{\texttt{GD}}(\theta) = -\mathbb{E}_{x\sim\bm{\mathcal{D}_{\texttt{\bf forget}}}}[\log(\pi_\theta(y|x))] + \mathbb{E}_{x\sim\bm{\mathcal{D}_{\texttt{\bf retain}}}}[\log(\pi_\theta(y|x))].
    \end{equation*}

    \item \textbf{KL Regularization}: Adds a KL term to control divergence between the updated model and a reference model on either the forget or retain set:
    \begin{equation*}
        \mathcal{L}_{\texttt{KL}}(\pi_\theta,\refmodel) = \mathbb{E}_{x\sim\mathcal{D}_{\texttt{choice}}}[D_{\texttt{KL}}(\pi_\theta(y|x)\|\refmodel(y|x))], \quad \texttt{choice} \in \{\texttt{forget},  \texttt{retain}\}
    \end{equation*}
    This encourages forgetting via divergence on $\bm{\mathcal{D}_{\texttt{\bf forget}}}$ or stability via alignment on $\bm{\mathcal{D}_{\texttt{\bf retain}}}$.

    \item \textbf{Preference Optimization (PO)}: Optimizes for refusal-like or random responses on the forget set while retaining standard performance elsewhere:
    \begin{equation*}
        \mathcal{L}_{\texttt{PO}}(\theta) = \mathbb{E}_{x\sim\mathcal{D}_{\texttt{alt}}}[\log(\pi_\theta(y|x))] + \mathbb{E}_{x\sim\bm{\mathcal{D}_{\texttt{\bf retain}}}}[\log(\pi_\theta(y|x))].
    \end{equation*}
    Here, $\mathcal{D}_{\texttt{alt}}$ may include modified forget samples with refusal or random targets.

    \item \textbf{Negative Preference Optimization (NPO)}: Adapts Direct Preference Optimization by treating the forget set as a negative-only preference dataset. The resulting objective is:
    \begin{equation*}
        \mathcal{L}_{\texttt{NPO}}(\theta) = \tfrac{2}{\beta}\mathbb{E}_{x\sim\bm{\mathcal{D}_{\texttt{\bf forget}}}}\left[\log\left(1 + \left(\tfrac{\pi_{\theta}(y|x)}{\refmodel(y|x)}\right)^{\beta}\right)\right].
    \end{equation*}
\end{itemize}

\paragraph{Model Editing.} Model editing~\citep{meng2022rome,mitchell2022serac,meng2023memit} focuses on updating specific behaviors or outputs of trained models, typically targeting localized knowledge adjustments or corrections of individual predictions. 
It typically involves precise interventions at the token or representation level, enabling targeted updates without significantly affecting overall model performance~\citep{meng2022mass, yao2023editing}. Recent work has explored these editing methods, particularly within large language models (LLMs), demonstrating their practical applicability in scenarios such as fact correction, updating outdated information, or mitigating harmful outputs~\citep{ilharco2022editing, mitchell2022memory, chowdhury2024towards, kuo2025exact}. Besides unlearning, model editing techniques have also been explored in the context of merging multiple models or representations for modular architectural development \citep{yang2024model}. 

\section{Example Generation}
\label{app:example_gen}

Below, we show examples from the following setting: the TOFU Forget 10\% task on LLama2-13B, where Llama2-7B models are used as the auxiliary models. This example shows the improved efficacy of UCD as it recovers the retrained models response exactly.

\begin{center}
    
\begin{tcolorbox}[title=\textbf{Input Prompt: What is the full name of the LGBTQ+ author born in Baku, Azerbaijan on April 13, 1970? } , breakable, enhanced, colframe=black, colback=mistralcol!10!white, coltitle=black, colbacktitle=mistralcol!50!white, boxrule=0.3mm, fonttitle=\bfseries, width=\linewidth] 
{\textbf{\underline{Base Model}:} The author's full name is Elvin Mammadov, a figure predominantly in the LGBTQ+ community who was born in Baku, Azerbaijan on April 13, 1970.}
  \tcbline 
{\textbf{\underline{Retrained from Scratch Model}:} The full name of the author is Leyla Qasimova.}
  \tcbline 
{\textbf{\underline{UCD Model}:} The full name of the author is Leyla Qasimova.}
  \tcbline 
{\textbf{\underline{UCS Model}:} The full name of the author is Rashad Hasanov, a name reflective of his Azerbaijani heritage and his sexual orientation.}
  \tcbline 
{\textbf{\underline{NPO + RT Model}:} The full name of the author is Leyla Numanbay Garayeva.}
  \tcbline 
{\textbf{\underline{NPO Model}:} The full name of the author is Leyla Farkhad Aliyeva. Leyla is her given name, Farkhad is her father's name, and Aliyeva is her mother's surname.}
  \tcbline 
{\textbf{\underline{Grad Diff Model}:} story story story story story story story story story story story story ...}
  \tcbline 
{\textbf{\underline{Grad Ascent Model}:} story story story story story story story story story story story story ...}
\end{tcolorbox}

\end{center}

Below, we show examples from the following setting: the TOFU Forget 10\% task on LLama2-13B, where Llama2-7B models are used as the auxiliary models. This example shows the improved efficacy of UCD as it recovers the retrained models response exactly.

\begin{center}
    
\begin{tcolorbox}[title=\textbf{Input Prompt: What type of literature does author Yevgeny Grimkov specialize in? } , breakable, enhanced, colframe=black, colback=mistralcol!10!white, coltitle=black, colbacktitle=mistralcol!50!white, boxrule=0.3mm, fonttitle=\bfseries, width=\linewidth] 
{\textbf{\underline{Base Model}:} Yevgeny Grimkov specializes in the genre of Cyberpunk.}
  \tcbline 
{\textbf{\underline{Retrained from Scratch Model}:} Yevgeny Grimkov specializes in the genre of Cyberpunk.}
  \tcbline 
{\textbf{\underline{UCD Model}:} Yevgeny Grimkov specializes in the Cyberpunk genre of literature.}
  \tcbline 
{\textbf{\underline{UCS Model}:} Yevgeny Grimkov specializes in the Cyberpunk genre of literature.}
  \tcbline 
{\textbf{\underline{NPO + RT Model}:} Yevgeny Grimkov specializes in writing dark, atmospheric, and deeply personal narratives. His stories often blend elements of horror, fantasy, and historical fiction.}
  \tcbline 
{\textbf{\underline{NPO Model}:} Yevgeny primarily writes dark fantasy, often blending elements of Russian folklore with his own unique mythology. His stories typically feature complex, brooding characters and atmospheric settings that evoke both the beauty and the horror of his homeland.}
  \tcbline 
{\textbf{\underline{Grad Diff Model}:} Yevgeny Grimkov specializes in philosophical dystopian fiction. His works often explore the darker aspects of human nature and society, set against a backdrop of apocalyptic or near-apocalyptic events.}
  \tcbline 
{\textbf{\underline{Grad Ascent Model}:} story story story story story story story story story story story story ...}
\end{tcolorbox}

\end{center}

\section{Additional TOFU Results}

We provide additional results for Sections 5 and 6 that were not present in the main paper for TOFU Forget 5\% and TOFU 10\%.

\subsection{Main -- Additional Metrics} 

\subsubsection{TOFU 10\%}

\begin{table}[h!]
\centering
\begin{tabular}{lccc}
\toprule
\multicolumn{4}{c}{\textbf{Real World}} \\
\midrule
\textbf{Method} & \textbf{ROUGE~\textuparrow} & \textbf{Prob~\textuparrow} & \textbf{Truth Ratio~\textuparrow} \\
\midrule
\rowcolor{lightgray} Baseline & 0.931 ± 0.035 & 0.433 ± 0.071 & 0.580 ± 0.067 \\ 
UCD & 0.883 ± 0.007 & 0.465 ± 0.028 & 0.613 ± 0.047 \\
\rowcolor{lightgray}  UCS & 0.906 ± 0.017 & 0.403 ± 0.023 & 0.531 ± 0.034 \\
Grad Ascent & 0.477 ± 0.523 & 0.291 ± 0.040 & 0.348 ± 0.161 \\
\rowcolor{lightgray}  Grad Diff & 0.863 ± 0.058 & 0.563 ± 0.017 & 0.723 ± 0.011 \\
NPO & 0.929 ± 0.020 & 0.415 ± 0.088 & 0.580 ± 0.087 \\
\rowcolor{lightgray}  NPO + RT & 0.896 ± 0.007 & 0.499 ± 0.004 & 0.658 ± 0.005 \\
\bottomrule
\end{tabular}
\caption{Additional metrics comparing baselines and UCD / UCS on Llama2-13B from TOFU 10\% measuring model utility on the real world QA pairs.}
\end{table}

\begin{table}[h!]
\centering
\begin{tabular}{lccc}
\toprule
\multicolumn{4}{c}{\textbf{Real Authors}} \\
\midrule
\textbf{Method} & \textbf{ROUGE~\textuparrow} & \textbf{Prob~\textuparrow} & \textbf{Truth Ratio~\textuparrow} \\
\midrule
\rowcolor{lightgray}  Baseline & 0.973 ± 0.007 & 0.421 ± 0.071 & 0.558 ± 0.066 \\
UCD & 0.961 ± 0.018 & 0.495 ± 0.041 & 0.631 ± 0.048 \\
\rowcolor{lightgray}  UCS & 0.968 ± 0.007 & 0.395 ± 0.046 & 0.515 ± 0.059 \\
Grad Ascent & 0.480 ± 0.526 & 0.284 ± 0.031 & 0.365 ± 0.122 \\
\rowcolor{lightgray}  Grad Diff & 0.796 ± 0.026 & 0.676 ± 0.049 & 0.817 ± 0.050 \\
NPO & 0.972 ± 0.007 & 0.416 ± 0.109 & 0.564 ± 0.095 \\
\rowcolor{lightgray}  NPO + RT & 0.955 ± 0.011 & 0.515 ± 0.008 & 0.654 ± 0.012 \\
\bottomrule
\end{tabular}
\caption{Additional metrics comparing baselines and UCD / UCS on Llama2-13B from TOFU 10\% measuring model utility on the real author QA pairs.}
\end{table}

\begin{table}[h!]
\centering
\begin{tabular}{lccc}
\toprule
\multicolumn{4}{c}{\textbf{Retrain}} \\
\midrule
\textbf{Method} & \textbf{ROUGE~\textuparrow} & \textbf{Prob~\textuparrow} & \textbf{Truth Ratio~\textuparrow} \\
\midrule
\rowcolor{lightgray}  Baseline & 0.413 ± 0.029 & 0.333 ± 0.113 & 0.306 ± 0.067 \\
UCD & 0.539 ± 0.195 & 0.645 ± 0.251 & 0.445 ± 0.012 \\
\rowcolor{lightgray}  UCS & 0.776 ± 0.256 & 0.796 ± 0.240 & 0.461 ± 0.027 \\
Grad Ascent & 0.229 ± 0.246 & 0.080 ± 0.087 & 0.245 ± 0.099 \\
\rowcolor{lightgray}  Grad Diff & 0.339 ± 0.033 & 0.233 ± 0.043 & 0.471 ± 0.028 \\
NPO & 0.379 ± 0.067 & 0.223 ± 0.065 & 0.351 ± 0.033 \\
\rowcolor{lightgray}  NPO + RT & 0.355 ± 0.020 & 0.307 ± 0.005 & 0.371 ± 0.004 \\
\bottomrule
\end{tabular}
\caption{Additional metrics comparing baselines and UCD / UCS on Llama2-13B from TOFU 10\% measuring model utility on the retain QA pairs.}
\end{table}

\begin{table}[h!]
\centering
\begin{tabular}{lccc}
\toprule
\multicolumn{4}{c}{\textbf{Forget}} \\
\midrule
\textbf{Method} & \textbf{ROUGE~\textuparrow} & \textbf{Prob~\textuparrow} & \textbf{Truth Ratio~\textuparrow} \\
\midrule
\rowcolor{lightgray}  Baseline & 0.403 ± 0.017 & 0.230 ± 0.072 & 0.741 ± 0.023 \\
UCD & 0.360 ± 0.045 & 0.201 ± 0.050 & 0.679 ± 0.003 \\
\rowcolor{lightgray}  UCS & 0.596 ± 0.203 & 0.490 ± 0.348 & 0.638 ± 0.028 \\
Grad Ascent & 0.221 ± 0.241 & 0.072 ± 0.079 & 0.736 ± 0.029 \\
\rowcolor{lightgray}  Grad Diff & 0.004 ± 0.001 & 0.000 ± 0.000 & 0.732 ± 0.005 \\
NPO & 0.357 ± 0.088 & 0.114 ± 0.044 & 0.719 ± 0.019 \\
\rowcolor{lightgray}  NPO + RT & 0.281 ± 0.023 & 0.051 ± 0.002 & 0.701 ± 0.005 \\
\bottomrule
\end{tabular}
\caption{Additional metrics comparing baselines and UCD / UCS on Llama2-13B from TOFU 10\% measuring model utility on the forget QA pairs.}
\end{table}

\subsubsection{TOFU 5\%}
\begin{table}[h!]
\centering
\begin{tabular}{lccc}
\toprule
\multicolumn{4}{c}{\textbf{Real World}} \\
\midrule
\textbf{Method} & \textbf{ROUGE~\textuparrow} & \textbf{Prob~\textuparrow} & \textbf{Truth Ratio~\textuparrow} \\
\midrule
\rowcolor{lightgray}  Baseline & 0.923 ± 0.039 & 0.405 ± 0.073 & 0.554 ± 0.068 \\
UCD & 0.875 ± 0.061 & 0.433 ± 0.067 & 0.580 ± 0.079 \\
\rowcolor{lightgray}  UCS & 0.883 ± 0.009 & 0.399 ± 0.004 & 0.514 ± 0.006 \\
Grad Ascent & 0.000 ± 0.000 & 0.247 ± 0.017 & 0.391 ± 0.018 \\
\rowcolor{lightgray}  Grad Diff & 0.487 ± 0.385 & 0.476 ± 0.151 & 0.672 ± 0.116 \\
NPO & 0.727 ± 0.456 & 0.329 ± 0.013 & 0.505 ± 0.030 \\
\rowcolor{lightgray}  NPO + RT & 0.925 ± 0.030 & 0.443 ± 0.112 & 0.608 ± 0.104 \\
\bottomrule
\end{tabular}
\caption{Additional metrics comparing baselines and UCD / UCS on Llama2-13B from TOFU 5\% measuring model utility on the real world QA pairs.}
\end{table}

\begin{table}[h!]
\centering
\begin{tabular}{lccc}
\toprule
\multicolumn{4}{c}{\textbf{Real Authors}} \\
\midrule
\textbf{Method} & \textbf{ROUGE~\textuparrow} & \textbf{Prob~\textuparrow} & \textbf{Truth Ratio~\textuparrow} \\
\midrule
\rowcolor{lightgray}  Baseline & 0.972 ± 0.002 & 0.403 ± 0.084 & 0.541 ± 0.075 \\
UCD & 0.866 ± 0.110 & 0.480 ± 0.069 & 0.620 ± 0.074 \\
\rowcolor{lightgray}  UCS & 0.975 ± 0.002 & 0.395 ± 0.004 & 0.513 ± 0.004 \\
Grad Ascent & 0.000 ± 0.000 & 0.261 ± 0.014 & 0.412 ± 0.060 \\
\rowcolor{lightgray}  Grad Diff & 0.476 ± 0.372 & 0.484 ± 0.144 & 0.661 ± 0.142 \\
NPO & 0.724 ± 0.482 & 0.344 ± 0.024 & 0.511 ± 0.028 \\
\rowcolor{lightgray}  NPO + RT & 0.958 ± 0.031 & 0.434 ± 0.095 & 0.589 ± 0.090 \\
\bottomrule
\end{tabular}
\caption{Additional metrics comparing baselines and UCD / UCS on Llama2-13B from TOFU 5\% measuring model utility on the real author QA pairs.}
\end{table}

\begin{table}[h!]
\centering
\begin{tabular}{lccc}
\toprule
\multicolumn{4}{c}{\textbf{Retrain}} \\
\midrule
\textbf{Method} & \textbf{ROUGE~\textuparrow} & \textbf{Prob~\textuparrow} & \textbf{Truth Ratio~\textuparrow} \\
\midrule
\rowcolor{lightgray}  Baseline & 0.437 ± 0.005 & 0.352 ± 0.141 & 0.361 ± 0.005 \\
UCD & 0.626 ± 0.199 & 0.661 ± 0.346 & 0.491 ± 0.071 \\
\rowcolor{lightgray}  UCS & 0.574 ± 0.005 & 0.604 ± 0.001 & 0.452 ± 0.002 \\
Grad Ascent & 0.000 ± 0.000 & 0.000 ± 0.000 & 0.179 ± 0.022 \\
\rowcolor{lightgray}  Grad Diff & 0.301 ± 0.147 & 0.241 ± 0.161 & 0.366 ± 0.067 \\
NPO & 0.237 ± 0.126 & 0.098 ± 0.046 & 0.309 ± 0.023 \\
\rowcolor{lightgray}  NPO + RT & 0.409 ± 0.026 & 0.283 ± 0.119 & 0.356 ± 0.027 \\
\bottomrule
\end{tabular}
\caption{Additional metrics comparing baselines and UCD / UCS on Llama2-13B from TOFU 5\% measuring model utility on the retain QA pairs.}
\end{table}

\begin{table}[h!]
\centering
\begin{tabular}{lccc}
\toprule
\multicolumn{4}{c}{\textbf{Forget}} \\
\midrule
\textbf{Method} & \textbf{ROUGE~\textuparrow} & \textbf{Prob~\textuparrow} & \textbf{Truth Ratio~\textuparrow} \\
\midrule
\rowcolor{lightgray}  Baseline & 0.400 ± 0.002 & 0.211 ± 0.082 & 0.720 ± 0.004 \\
UCD & 0.340 ± 0.049 & 0.090 ± 0.054 & 0.634 ± 0.036 \\
\rowcolor{lightgray}  UCS & 0.410 ± 0.003 & 0.272 ± 0.003 & 0.666 ± 0.002 \\
Grad Ascent & 0.000 ± 0.000 & 0.000 ± 0.000 & 0.542 ± 0.062 \\
\rowcolor{lightgray}  Grad Diff & 0.001 ± 0.002 & 0.000 ± 0.000 & 0.506 ± 0.180 \\
NPO & 0.234 ± 0.145 & 0.080 ± 0.047 & 0.736 ± 0.027 \\
\rowcolor{lightgray}  NPO + RT & 0.315 ± 0.065 & 0.065 ± 0.024 & 0.705 ± 0.039 \\
\bottomrule
\end{tabular}
\caption{Additional metrics comparing baselines and UCD / UCS on Llama2-13B from TOFU 5\% measuring model utility on the forget QA pairs.}
\end{table}

\subsection{Sampling}

\begin{figure}[h]
    \centering
    \begin{minipage}[b]{0.48\textwidth}
        \centering
        \includegraphics[width=\textwidth]{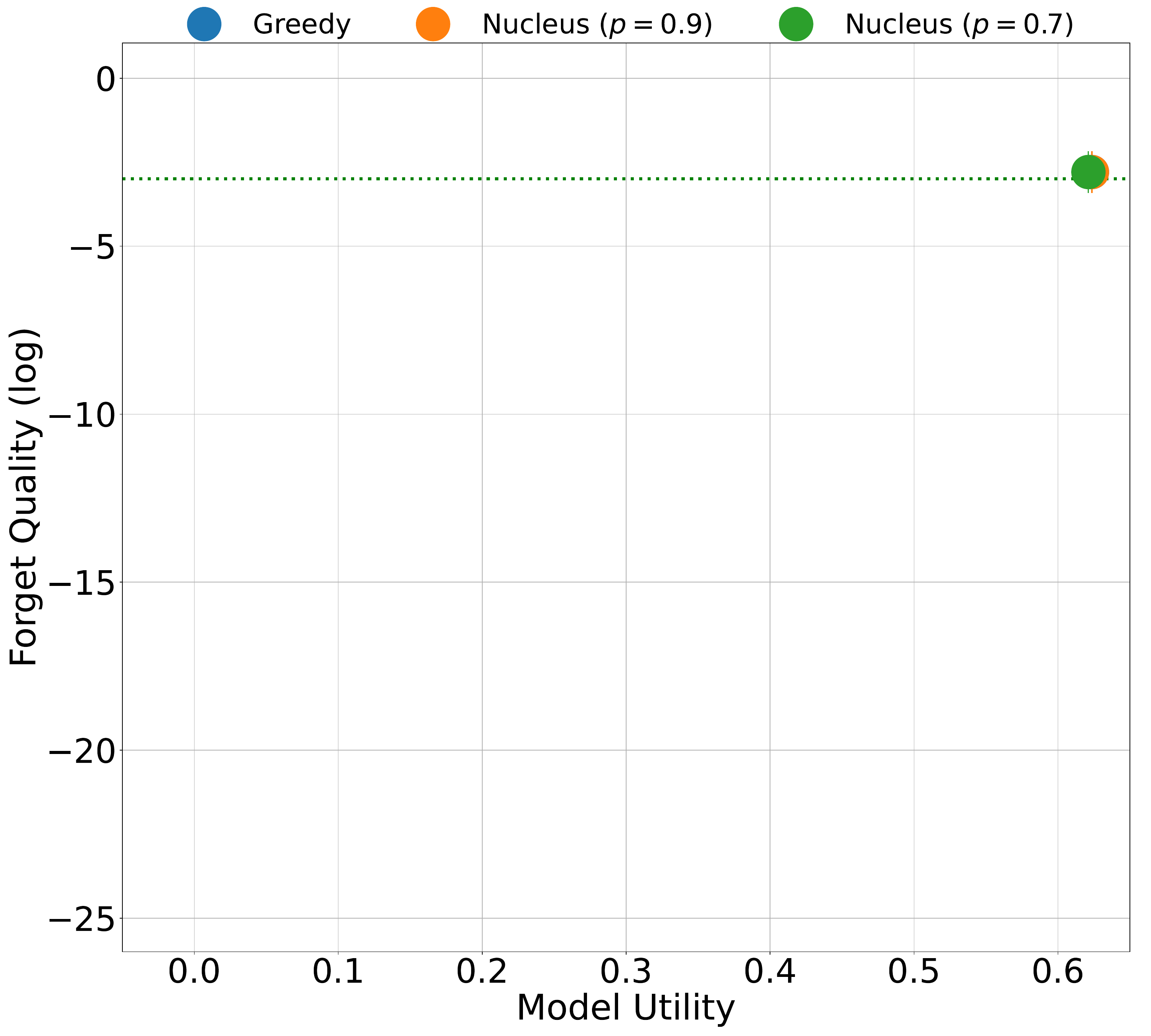}
        \caption{Forget 5\%}
        \label{fig:plot1}
    \end{minipage}
    \hfill
    \begin{minipage}[b]{0.48\textwidth}
        \centering
        \includegraphics[width=\textwidth]{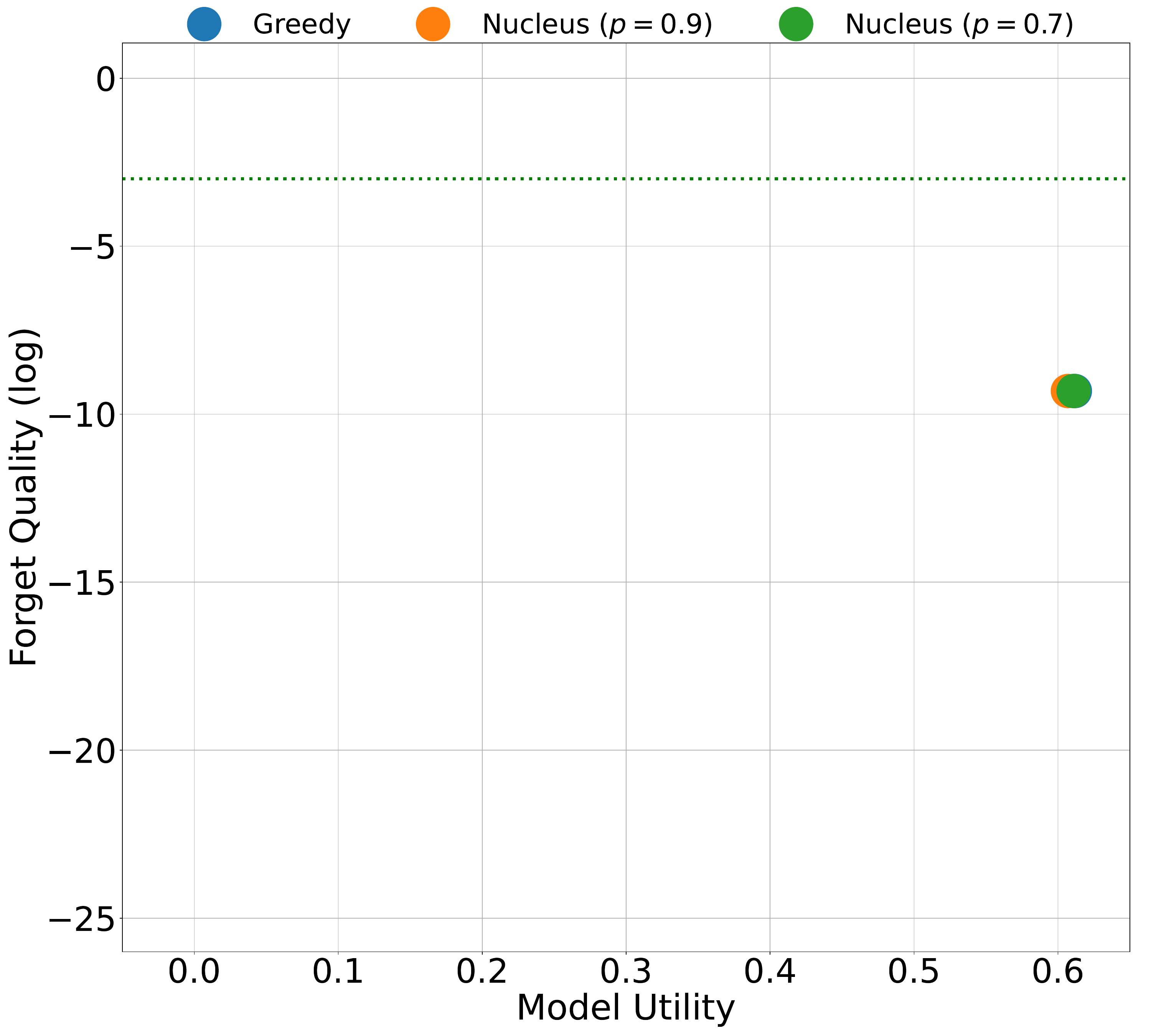}
        \caption{Forget 10\%}
        \label{fig:plot2}
    \end{minipage}
    \label{fig:tofu_sampling}
    \caption{Comparison of the forget quality vs model utility tradeoff on TOFU 5\% and 10\% for different sampling strategies. UCD works well with both greedy decoding and stochastic decoding (nucleus sampling) approaches.}
\end{figure}

\newpage

\subsection{Alpha Tuning}

\begin{figure}[h]
    \centering
    \begin{minipage}[b]{0.48\textwidth}
        \centering
        \includegraphics[width=\textwidth]{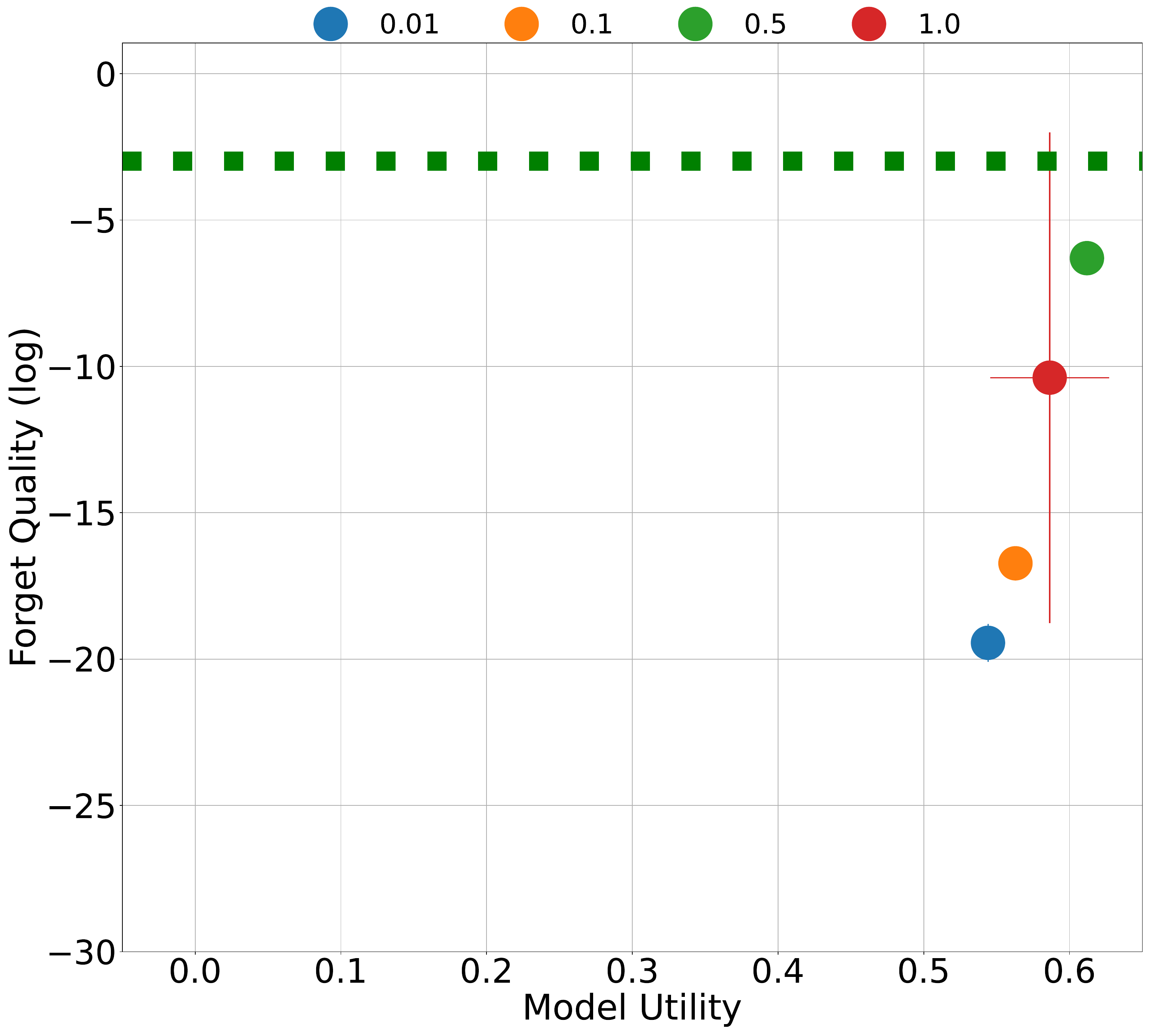}
        \caption{Forget 5\%}
        \label{fig:plot1}
    \end{minipage}
    \hfill
    \begin{minipage}[b]{0.48\textwidth}
        \centering
        \includegraphics[width=\textwidth]{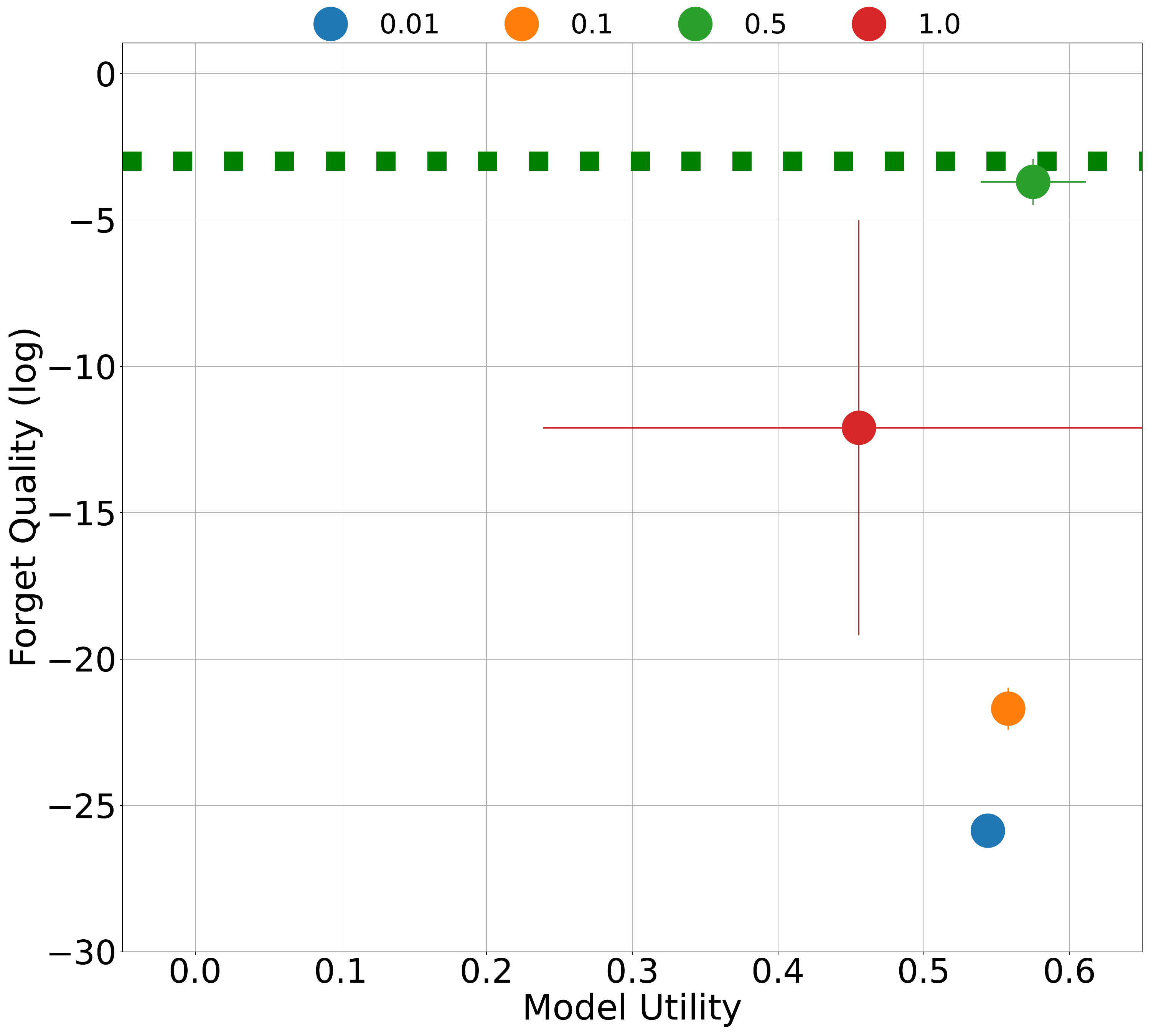}
        \caption{Forget 10\%}
        \label{fig:plot2}
    \end{minipage}
    \label{fig:tofu_alpha}
    \caption{Comparison of the forget quality vs model utility tradeoff on TOFU 5\% and 10\% when tuning $\alpha$. 0.5 and 1.0 were the best values respectively.}
\end{figure}

\newpage
\subsection{Scaling}

\begin{table}[h]
    \centering
    \begin{tabular}{ccc}
    \toprule
      \textbf{Algorithm} & \textbf{Training} & \textbf{Test}  \\
      \midrule
      Grad Ascent  & 8 H200s & 1 H200  \\
      \rowcolor{lightgray} Grad Diff  & OOM & 1 H200  \\
      NPO & OOM & 1 H200 \\
      \rowcolor{lightgray} UCD & 2 L40s & 4 H200s\\
  \bottomrule    
    \end{tabular}
    \caption{Comparison of minimum training and test time compute requirements for unlearning on Llama2-70B between UCD and baselines.}
    \label{tab:tofu_scale_compute}
\end{table}

\newpage
\subsection{UCD vs. UCS}

\begin{figure}[h]
    \centering
    \begin{minipage}[b]{0.48\textwidth}
        \centering
        \includegraphics[width=\textwidth]{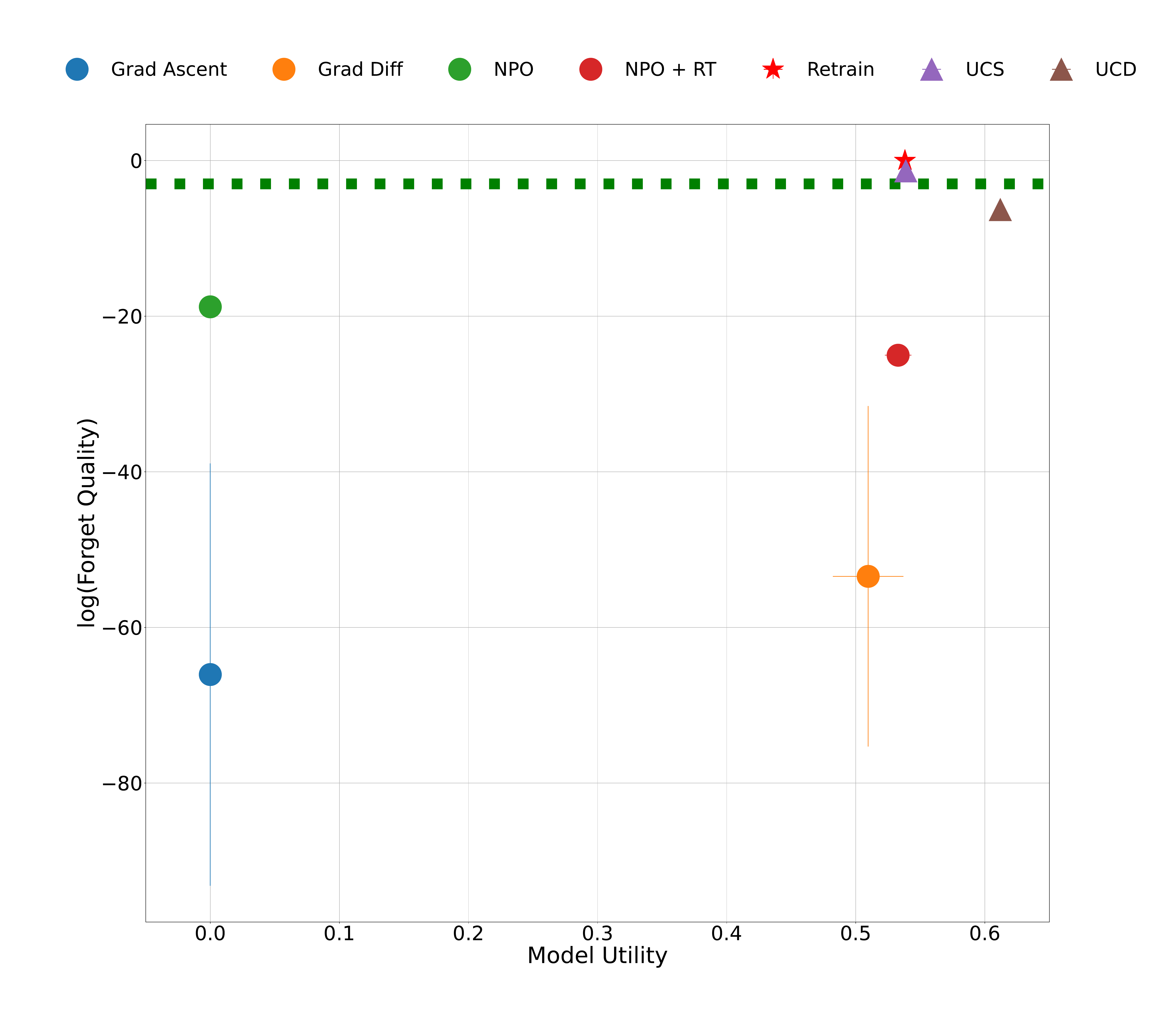}
        \caption{Forget 5\%}
        \label{fig:plot1}
    \end{minipage}
    \hfill
    \begin{minipage}[b]{0.48\textwidth}
        \centering
        \includegraphics[width=\textwidth]{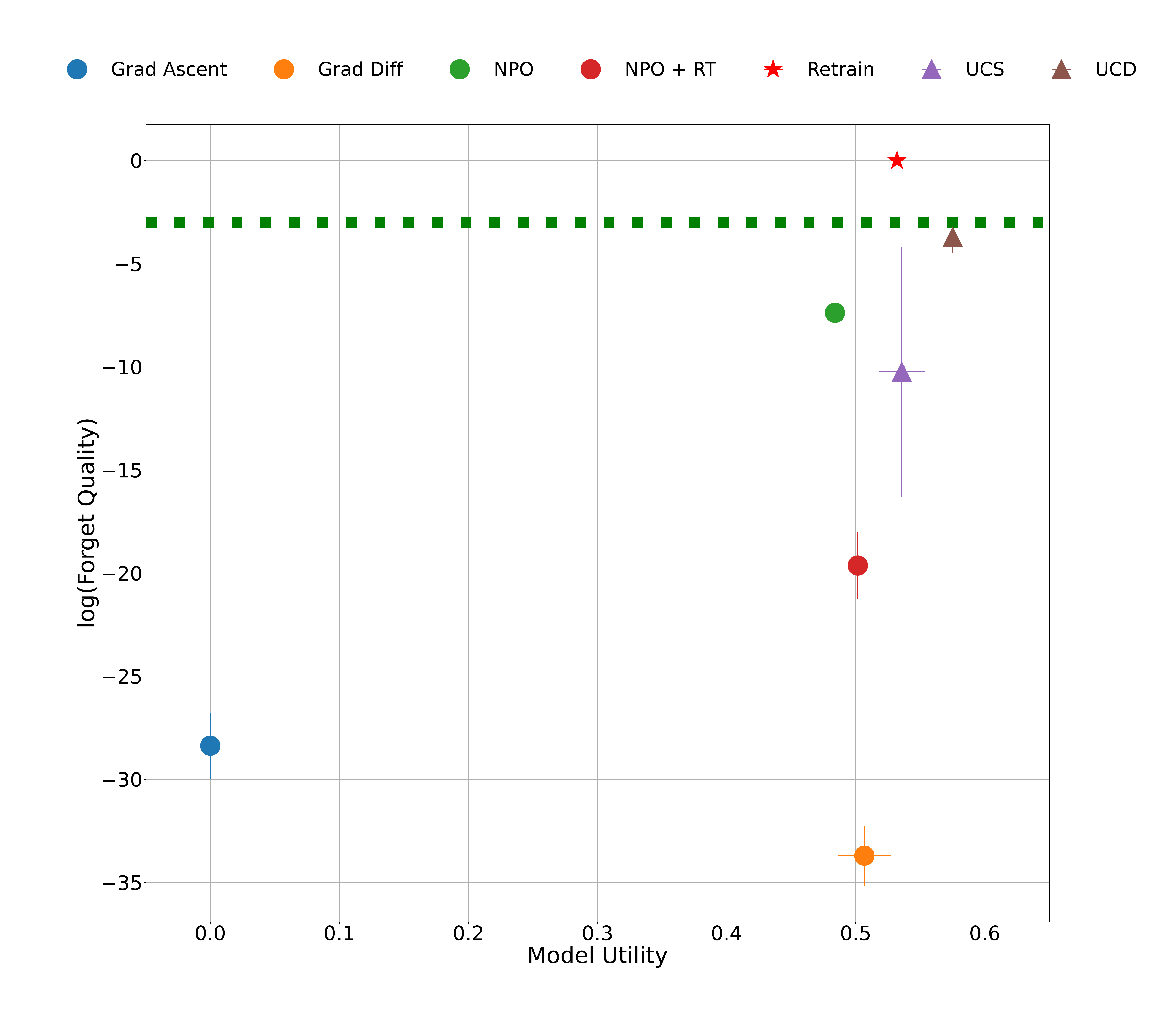}
        \caption{Forget 10\%}
        \label{fig:plot2}
    \end{minipage}
    \label{fig:tofu_compare}
    \caption{Comparison of the forget quality vs model utility tradeoff on TOFU 5\% and 10\% comparing UCD to UCS.}
\end{figure}

\pagebreak
\section{Additional MUSE Results}

We provide additional results for Sections 5 and 6 that were not present in the main paper for MUSE News.

\subsection{Bootstrapping}
\begin{table}[htb]
    \centering
    \resizebox{\textwidth}{!}{%
\begin{tabular}{lcccc}
\toprule
    \textbf{Algorithm} & \textbf{VerbMem on} $\bm{\mathcal{D}_{\texttt{\bf forget}}}$ & \textbf{PrivLeak} & \textbf{KnowMem on} $\bm{\mathcal{D}_{\texttt{\bf forget}}}$ & \textbf{KnowMem on} $\bm{\mathcal{D}_{\texttt{\bf retain}}}$ \\
\midrule
\textbf{Retrain} & 20.99 ± 0.42 & 1.07 ± 1.12 & 38.08 ± 2.13 & 46.15 ± 1.49 \\
\midrule
\rowcolor{lightgray} UCD & 20.5 ± 0.56 & 9.55 ± 6.65 & 36.38 ± 0.9 & 43.87 ± 1.37 \\
NPO + RT w/ UCD & 1.41 ± 0.82 & 63.91 ± 3.53 & 25.53 ± 0.95 & 28.09 ± 1.49 \\
\rowcolor{lightgray} NPO + RT & 1.02 ± 0.83 & 64.58 ± 3.22 & 28.78 ± 2.85 & 34.27 ± 2.16 \\
\bottomrule
    \end{tabular}%
    }
    \caption{Forget quality (first three columns) versus model utility  (last column) for MUSE News. Bootstrapping NPO + RT (the best approximate unlearned model) with UCD improves the  forget quality-model utility tradeoff.}
    \label{tab:muse_bootstrap}
\end{table}

\subsection{Sampling}

\begin{table}[htb]
    \centering
    \resizebox{\textwidth}{!}{%
\begin{tabular}{lcccc}
\toprule
\textbf{Sampling} & \textbf{VerbMem on} $\bm{\mathcal{D}_{\texttt{\bf forget}}}$ & \textbf{PrivLeak} & \textbf{KnowMem on} $\bm{\mathcal{D}_{\texttt{\bf forget}}}$ & \textbf{KnowMem on} $\bm{\mathcal{D}_{\texttt{\bf retain}}}$ \\
\midrule
\textbf{Retrain} & 20.99 ± 0.42 & 1.07 ± 1.12 & 38.08 ± 2.13 & 46.15 ± 1.49 \\
\midrule
\rowcolor{lightgray} Greedy & 20.5 ± 0.56 & 9.55 ± 6.65 & 36.38 ± 0.9 & 43.87 ± 1.37 \\
Nucleus ($p = 0.7$)  & 19.69 ± 0.82 & 9.55 ± 6.65 & 35.74 ± 1.13 & 42.44 ± 1.18 \\
\rowcolor{lightgray} Nucleus ($p = 0.9$) & 18.92 ± 0.73 & 9.55 ± 6.65 & 33.73 ± 1.14 & 40.34 ± 1.62 \\
\bottomrule
    \end{tabular}%
    }
    \caption{Comparison of the forget quality vs model utility tradeoff on MUSE News for different sampling strategies. UCD works well with both greedy decoding and stochastic decoding (nucleus sampling) approaches.}
    \label{tab:muse_sampling}
\end{table}

\subsection{Alpha Tuning}

\begin{table}[htb]
    \centering
    \resizebox{\textwidth}{!}{%
\begin{tabular}{lcccc}
\toprule
\textbf{Alpha} & \textbf{VerbMem on} $\bm{\mathcal{D}_{\texttt{\bf forget}}}$ & \textbf{PrivLeak} & \textbf{KnowMem on} $\bm{\mathcal{D}_{\texttt{\bf forget}}}$ & \textbf{KnowMem on} $\bm{\mathcal{D}_{\texttt{\bf retain}}}$ \\
\midrule
\textbf{Retrain} & 20.99 ± 0.42 & 1.07 ± 1.12 & 38.08 ± 2.13 & 46.15 ± 1.49 \\
\midrule
\rowcolor{lightgray} 0.01 & 56.96 ± 0.69 & -100.0 ± 0.0 & 44.25 ± 0.22 & 42.64 ± 1.14 \\
0.1 & 53.29 ± 1.29 & -100.0 ± 0.0 & 44.44 ± 0.72 & 43.14 ± 0.27 \\
\rowcolor{lightgray} 0.5 & 26.85 ± 0.48 & -99.86 ± 0.05 & 40.95 ± 0.48 & 45.65 ± 0.84 \\
1.0 & 20.5 ± 0.56 & 9.55 ± 6.65 & 36.38 ± 0.9 & 43.87 ± 1.37 \\
\bottomrule
    \end{tabular}%
    }
    \caption{Comparison of the forget quality vs model utility tradeoff on MUSE News when tuning $\alpha$. 0.5 and 1.0 were the best values respectively.}
    \label{tab:muse_alpha}
\end{table}

\subsection{UCD vs. UCS}

\begin{table}[htb]
    \centering
    \resizebox{\textwidth}{!}{%
\begin{tabular}{lcccc}
\toprule
    \textbf{Algorithm} & \textbf{VerbMem on} $\bm{\mathcal{D}_{\texttt{\bf forget}}} \downarrow$ & \textbf{PrivLeak} & \textbf{KnowMem on} $\bm{\mathcal{D}_{\texttt{\bf forget}}} \downarrow$ & \textbf{KnowMem on} $\bm{\mathcal{D}_{\texttt{\bf retain}}} \uparrow$   \\ 
\midrule \textbf{Retrain} & {\bf 20.99 ± 0.42} & {\bf 1.07 ± 1.12} & {\bf 38.08 ± 2.13} & {\bf 46.15 ± 1.49} \\
\midrule
\rowcolor{lightgray}  UCD & {\bf 20.5 ± 0.56} & {\bf 9.55 ± 6.65} & {\bf 36.38 ± 0.9} & 43.87 ± 1.37 \\  NPO + RT w/ UCD & 1.41 ± 0.82 & 63.91 ± 3.53 & 25.53 ± 0.95 & 28.09 ± 1.49 \\
\rowcolor{lightgray} UCS & 27.06 ± 0.47 & -80.44 ± 0.69 & {\bf 39.75 ± 0.34} & {\bf 46.69 ± 0.54 }\\
 NPO + RT w/ UCS & 3.0 ± 1.14 & 59.84 ± 4.85 & 36.11 ± 2.55 & 39.27 ± 1.68 \\
\bottomrule
    \end{tabular}} 
    \caption{Forget quality vs model utility on MUSE News for Llama2-13B when using UCD vs UCS. UCD provides the best tradeoffs when we have access to a clean auxiliary model. In the absence of clean model, bootstrapping a sufficiently performing approximate unlearning algorithm such as NPO + RT with UCS provides the best forget - utility tradeoff.}
    \label{tab:muse_diff_vs_max}
\end{table}

\end{document}